\pdfoutput=1
\documentclass[10pt, a4paper, twocolumn, showabstract]{naverlabseurope}

\usepackage{lipsum}
\usepackage{multicol}
\usepackage{tikz,tkz-kiviat,pgfplots}
\usepackage{times}
\usepackage{latexsym}

\usepackage{geometry}
\usepackage{tabularx}
\usepackage{makecell} % Required for \makecell
\usepackage{multirow}
\usepackage{float}     % For figure positioning
\usepackage{amsmath}
\usepackage{pifont}
\usepackage{hyperref}
\usepackage{verbatim}
\usepackage{booktabs}
\usepackage{tcolorbox}
\usepackage{xcolor}   % For coloring text
\usepackage{enumitem}
\definecolor{algocolor1}{rgb}{0.400, 0.761, 0.647}
\definecolor{algocolor2}{rgb}{0.988, 0.553, 0.384}
\definecolor{algocolor3}{rgb}{0.553, 0.627, 0.796}
\definecolor{algocolor4}{rgb}{0.906, 0.541, 0.765}
\definecolor{algocolor5}{rgb}{0.651, 0.847, 0.329}
\definecolor{algocolor6}{rgb}{1.000, 0.851, 0.184}
\definecolor{algocolor7}{rgb}{0.898, 0.769, 0.580}
\definecolor{algocolor0}{rgb}{0.702, 0.702, 0.702}

\newcommand{\cmark}{\textcolor{green!70!black}{\scalebox{1.2}{\ding{51}}}} % Green checkmark
\newcommand{\xmark}{\textcolor{red!70!black}{\scalebox{1.2}{\ding{55}}}}   % Red crossmark

\usepackage[T1]{fontenc}
\usepackage[utf8]{inputenc}

\usepackage{microtype}

\usepackage{inconsolata}

\usepackage{graphicx}

\title{PISCO: Pretty Simple Compression for Retrieval-Augmented Generation}
\titlerunning{PISCO}

\correspondingauthor{maxime.louis@naverlabs.com}

\authors{Maxime Louis \authsep Hervé Déjean \authsep Stéphane Clinchant}
\affiliations{NAVER LABS Europe}
\contributions{}
\website{}
\websiteref{}

\begin{abstract}
Retrieval-Augmented Generation (RAG) pipelines enhance Large Language Models (LLMs) by retrieving relevant documents, but they face scalability issues due to high inference costs and limited context size. Document compression is a practical solution, but current soft compression methods suffer from accuracy losses and require extensive pretraining. In this paper, we introduce PISCO\footnote{Code and models will be released soon.}, a novel method that achieves a 16x compression rate with minimal accuracy loss (0-3\%) across diverse RAG-based question-answering (QA) tasks. Unlike existing approaches, PISCO requires no pretraining or annotated data, relying solely on sequence-level knowledge distillation from document-based questions. With the ability to fine-tune a 7-10B LLM in 48 hours on a single A100 GPU, PISCO offers a highly efficient and scalable solution. We present comprehensive experiments showing that PISCO outperforms existing compression models by 8\% in accuracy.
\end{abstract}

\begin{document}

\maketitle

\section{Introduction}

Retrieval-Augmented Generation (RAG) \cite{lewis2020retrieval, guu2020retrieval, borgeaud2022improving} pipelines have become a crucial component in addressing various natural language tasks. By incorporating documents retrieved from a selected collection, RAG enhances Large Language Models (LLMs) enabling them to provide more accurate, current, and domain-specific responses.

The primary drawback is the increased inference cost, which scales quadratically with the number of tokens and, consequently, with the number of retrieved documents when using transformer-based architectures. In addition to inference costs, the limitations on LLM context size restrict the number of documents—and thus the amount of information—that can be utilized. This constrains the potential scaling of inference time \cite{yue2024inference}.

\begin{figure}[t]
    \centering
    \includegraphics[width=0.9\columnwidth]{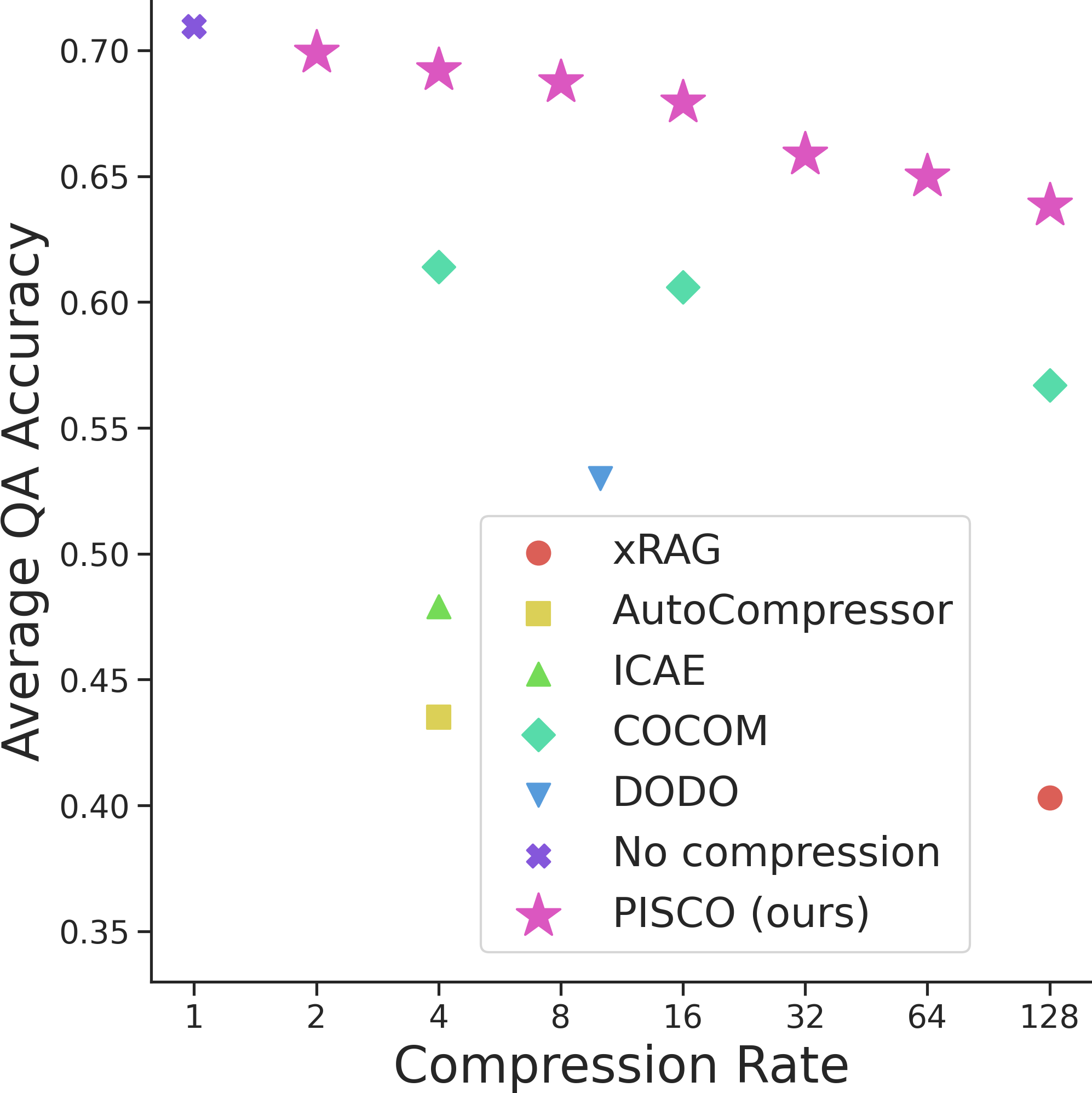} 
    \caption{PISCO substantially outperforms existing context compression methods for question answering with RAG. Shown here with Mistral-7B backbone.}
    \label{fig:intro_figure}
\end{figure}

Compressing documents is a practical way to reduce the computational burden of processing large contexts. Hard compression techniques focus on altering the surface structure of the documents, such as by pruning \cite{pan2024llmlingua, li2023compressing, wang2023learning, provence} or summarization \cite{xu2023recomp}. These methods are easily interpretable and can typically be applied to any LLM without requiring modifications. However, the compression rate is limited by the amount of information that can be effectively conveyed through text tokens, usually achieving a reduction of 2x-5x.

\begin{figure*}[!tb]
    \centering
    \includegraphics[width=\textwidth, trim=0 12 0 8, clip]{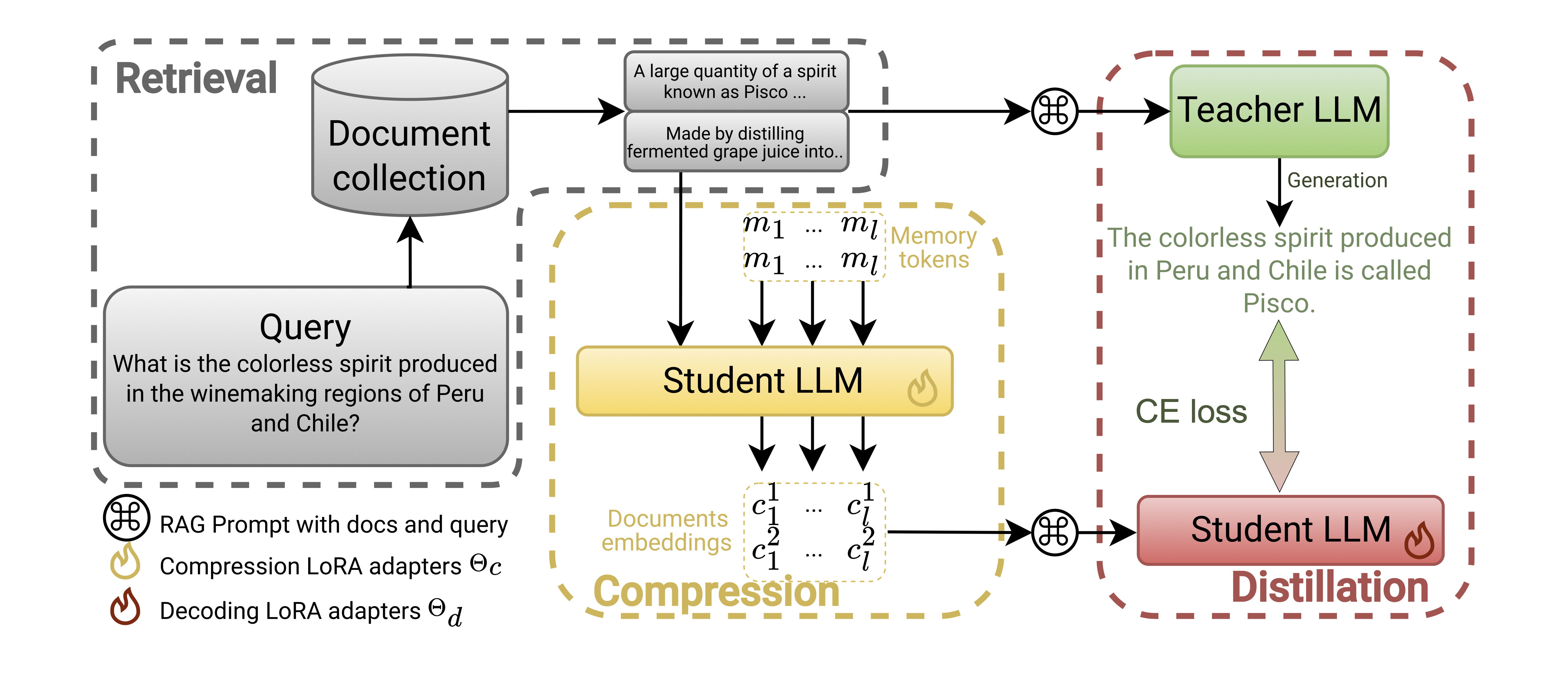} 
    \caption{Overview of PISCO training, shown here with $k=2$ documents. Training is supervised by distillation from a teacher model. Once trained, the full collection of documents can be compressed once to allow fast inference.}
    \label{fig:pisco_method}
\end{figure*}

Soft compression techniques aim to condense documents into vector representations \cite{wingate2022prompt}. They may also involve attention key-value pairs that the model attends to during generation, either through self-attention \cite{rau2024bergen, cheng2024xrag} or dedicated cross-attention mechanisms \cite{yen2024long}. These  methods trade off interpretability for efficiency, achieving higher compression rates while maintaining some performance levels. Most existing soft compression approaches for RAG follow a similar pipeline \cite{cheng2024xrag, chevalier2023adapting, ge2023context, rau2024context, qin2024dodo, li2024500xcompressor, yen2024long}. Typically, a pretraining task (such as auto-encoding and/or causal language modeling) on unlabeled data is used to train an initial compressor, followed by fine-tuning for RAG question answering (QA) to optimize the embeddings for QA tasks. 

Currently, all existing soft methods experience significant accuracy losses ( > 8\%, see Table \ref{table:main_results}) on RAG-QA benchmarks when compared to the original, uncompressed generator. This hinders the deployment of such systems, as accuracy is a primary concern over inference costs for most RAG systems. Additionally, all existing methods require pretraining on a large dataset as well as annotated QA datasets for fine-tuning.

This paper presents \textbf{PISCO, a compression method for RAG that achieves a x16 document compression rate with minimal to no loss in accuracy} (0-3\%) across a wide range of RAG-QA tasks, covering multiple domains. Unlike prior approaches, \textbf{PISCO requires neither pretraining nor annotated data}: it relies solely on distillation from open-ended, document-based questions. With x16 compression, PISCO models achieve a 5.7x inference speed-up. This makes PISCO a highly efficient and scalable solution for practical applications. Notably, \textbf{fine-tuning a 7-10B LLM into a PISCO model can be completed in under 48H on a single A100 GPU}. 

Here is a summary of our main contributions:
\begin{itemize}\setlength{\itemsep}{0pt}
    \item A simplified and more efficient pipeline for training compression models for RAG, %requiring no pretraining and no labeled data.
    \item Strong experiments results on in-domain, out-of-domain and multilingual QA: PISCO\footnote{Models are available at \href{https://huggingface.co/naver/pisco-mistral}{huggingface.co/naver/pisco-mistral}} out-performs current state-of-the-art compression models by 8\% in accuracy.
    \item An in-depth analysis demonstrating that pretraining has little benefits for compression models, investigating the quality of labels and illustrating the structure of the compressed documents embeddings.
\end{itemize}

In Section \ref{section:related_work}, we review related work on compression techniques for RAG. Section \ref{section:methods} describes the proposed method for PISCO. We then report experimental results on standard RAG benchmark \ref{subsection:main_results}. Furthermore, the importance of the design choices of PISCO is analyzed in section \ref{subsection:pisco_is_optimal}. Finally, we evaluate PISCO robustness and generalization to new tasks and domains in section \ref{subsection:out_of_domain}. 

%out-of-domain and multilingual data \S \ref{subsection:out_of_domain}  across various PISCO models with different backbones and configurations. Additionally, we illustrate in \S \ref{subsection:pisco_is_optimal} the importance of the design choices of PISCO: the impact of pre-training tasks and teacher choice on the final performances of compression models. %We further provide insights into how information is stored within the compressed document embeddings.

\section{Related Work}
\label{section:related_work}

\begin{table*}[!tb]
  \centering
  \resizebox{\textwidth}{!}{
    \begin{tabular}{|l|>{\centering\arraybackslash}p{2cm}|>{\centering\arraybackslash}p{2cm}|>{\centering\arraybackslash}p{2cm}|>{\centering\arraybackslash}p{2cm}|>{\centering\arraybackslash}p{2.5cm}|>{\centering\arraybackslash}p{2cm}|}
      \hline
      \multirow{2}{*}{\textbf{Paper}} & \textbf{Compression rate}$^\dagger$ & \multicolumn{2}{c|}{\textbf{Pre-training Required}} & \textbf{Distillation fine-tuning} & \textbf{Decoder training} & \textbf{Supervised fine-tuning} \\
      \cline{3-4}
      & & \makecell{\textbf{AE}} & \makecell{\textbf{LM}} & & & \\
      \hline
      AutoCompressor \cite{chevalier2023adapting} & 6-40x & \xmark & \cmark & \xmark & \cmark & \xmark \\
      \hline
      ICAE \cite{ge2023context} & 2-8x & \cmark & \cmark & \xmark & \xmark & \cmark \\      
      \hline
      xRAG \cite{cheng2024xrag} & 100-180x & \cmark & \cmark & \cmark$^\ddagger$ & \xmark & \cmark \\
      \hline
      DODO \cite{qin2024dodo} & 5-10x & \xmark & \cmark & \xmark & \cmark & \xmark \\
      \hline
      x500 \cite{li2024500xcompressor} & 6-480x & \cmark & \xmark & \xmark & \xmark & \cmark \\
      \hline
      CEPE \cite{yen2024long} & $\sim$ 256x & \xmark & \cmark & \cmark$^\ddagger$ & \cmark & \cmark \\
      \hline
      COCOM \cite{rau2024context} & 4-128x & \cmark & \cmark & \xmark & \cmark & \cmark \\
      \hline
      PISCO (ours) & 2-128x & \xmark & \xmark & \cmark & \cmark & \cmark \\
      \hline
    \end{tabular}
    }
    \caption{Summary of papers with different soft compression strategies. AE=Auto-encoding, LM=next token prediction. \textit{$^\dagger$ Compression rates are taken from QA/RAG experiments when available. $^\ddagger$ xRAG/CEPE use token-level distillation which requires ground truth labels.}}
    \label{table:related_work}
\end{table*}

Table \ref{table:related_work} compares key methods for soft compression, which, although not always explicitly designed for this purpose, can be applied in RAG (retrieval-augmented generation) applications. \cite{verma2024contextualcompression_survey} presents a more thorough survey or introduction to context compression.

\subsection{Dealing with long contexts via compression}

In \cite{chevalier2023adapting}, the authors present the Autocompressor, a recursive context compression method trained on a language modeling task. By appending compression tokens to the context and extracting the hidden states, this approach supports longer contexts and can be applied to document compression in RAG-QA. 
The in-context auto-encoder (ICAE) \cite{ge2023context} simplifies this by freezing the decoder, removing recursion, and pretraining through document auto-encoding. 
In \cite{yen2024long}, multiple contexts are encoded in parallel, with cross-attention layers introduced between the query and documents in the decoder LLM. This separation of query-document and self-attention reduces complexity and accelerates inference. However, large compressed documents remain a limitation for efficient storage. 
Finally, DODO \cite{qin2024dodo} compresses earlier context sections into adaptive \textit{nugget} states, using cross-attention for nuggets and self-attention for recent context. Though not specifically optimized for document QA, it can be used in that perspective.

\subsection{Compression specific to RAG-QA}

In \cite{rau2024bergen}, the authors specifically address the RAG-QA problem. After large-scale auto-encoding and language modeling pretraining, they fine-tune their decoder models to handle multiple documents simultaneously. Although this approach enhances the usability and performance of the RAG pipeline, there remains a significant performance drop (\textasciitilde 8\%) between uncompressed and x16 compressed models. The x500 Compressor \cite{li2024500xcompressor} is similar to COCOM except that the document embeddings consist directly of the K/V values obtained on the memory tokens during forward pass. This saves decoding computations but substantially increases the storage size of the embeddings. The xRAG method \cite{cheng2024xrag} proposes leveraging existing document embeddings—such as those used for retrieval—to reduce the storage and computational costs of generating additional embeddings. To achieve this, they train a small adapter to map the retrieval embeddings into the input space of a frozen decoder LLM. Similar to \cite{yen2024long}, xRAG also utilizes token-level distillation for fine-tuning in QA tasks.

All current compression methods \cite{cheng2024xrag, chevalier2023adapting, ge2023context, rau2024context, qin2024dodo, li2024500xcompressor, yen2024long} rely on large-scale pretraining tasks and require annotated labels. Despite their advancements, these methods still fall short of achieving the QA performance of their uncompressed LLM backbones (see \ref{table:main_results}).

% TODO: compare the size of QA tuning to explain

%For hard compression, mention adacomp \cite{zhang2024adacomp} which compresses the NUMBER of documents, query-dependently.

% REPLUG TODO
% AutoCompressor is not RAG specific and their RAG experiments only measure perplexity. For these RAG experiments the compression rate is 6.4 (but results are not great from their own confession. they compress up to x400 for text continuation
% Findings of the Association for Computational Linguistics: EMNLP 2022, pages 5621–5634December 7-11, 2022 ©2022 Association for Computational Linguistics Prompt Compression and Contrastive Conditioning for Controllability and Toxicity Reduction in Language Model
% About DODO: they train with one adapter for encoder, one adapter for decoder. Their RAG benchmark is weak (SQUAD, n_doc=1) with no real result, code/model is not available

% TODO: LMSUMM

% GritLM is not really compressive, but it saves computations.

% NB: Gist paper is in fact related to prompt compression i.e.  for the system prompt of GPT4, it's NOT comparable to COCOM in fact.

%Note: K/V information may be richer ! but it is much larger to store
% x500 compressor is available
% Should we include the number of params somewhere ? It coudl matter

\section{Methods}
\label{section:methods}

\subsection{Retrieval-Augmented Generation}

In RAG, each query $q$ is augmented with a set of relevant documents $(d_1, d_2,\ldots, d_k)$ retrieved from a large database of documents $\mathcal D$. For improved performance, this process typically involves two steps: first, a retriever identifies an initial pool of relevant documents, and then a re-ranker refines and prioritizes the $k$ most relevant ones. The final response $r$ is generated by prompting an language model $\mathcal F$ with both the query and the set of retrieved documents.

In general, the accuracy of the generated response improves as the number of documents increases. However, since documents tend to be longer than queries and the computational complexity of transformer-based models scales quadratically with the context length, this can make generation computationally expensive and cause delays. A soft compression model addresses this by mapping each document $d_i$ into a shorter set of embeddings or a key-value (K/V) cache, $\mathbf{c_i}$. The generation process is then conditioned on these compressed representations: $r \sim \mathcal{F}( \cdot\mid q, \mathbf{c_1}, \mathbf{c_2}, \dots, \mathbf{c_k})$.

\subsection{PISCO}
PISCO adopts a standard architecture, involving a compressor and decoder model, detailed in the following section. The main difference lies in its training task. The method is described on Figure \ref{fig:pisco_method}. 

\paragraph{Compression} is performed following the approach in \cite{chevalier2023adapting, ge2023context, rau2024context}, utilizing the language model $\mathcal{F}$ with LoRA \cite{hu2021lora} adapters $\theta_c$. Specifically, a set of $l$ memory tokens $(m_1,\ldots, m_l)$ is appended to each document $d_i$: the corresponding prompt $(d_i; m_1, \ldots, m_l)$ is passed through $\mathcal{F}_{\theta_c}$. The $l$ final hidden states, corresponding to the memory tokens, are extracted to form the document embeddings $\mathbf{c_i}=(c_i^s)_{s=1\ldots l}$. Each document is encoded into $l$ vectors, each sharing the dimension of the encoder’s embedding space, $\mathcal F$. The number of tokens $l$ effectively controls the compression rate. The memory tokens $(m_1,\ldots, m_l)$ are optimized jointly with the LoRA adapters $\theta_c$. During optimization, $\mathcal{F}_{\theta_c}$ is encouraged via the distillation objective described below to encode information about the compressed document.

\paragraph{Decoding} is carried out using the language model $\mathcal{F}$ with a separate set $\theta_d$ of LoRA adapters. Previous works \cite{cheng2024xrag, ge2023context, li2024500xcompressor} attempt to freeze the decoder, to allow for plug-and-play use of the compressor. Early experiments suggested such an approach is unlikely to reach satisfying results (see Appendix \ref{appendix:frozen_decoder}). Fine-tuning the decoder might be crucial as it allows to adapt its interactions with the compressed representations \textbf{depending on the query}. Additionally, this fine-tuning process does not compromise the ease of setting up PISCO, as our end-to-end fine-tuning completes in only a day for $7-10$B-parameter models on a single high-end GPU.

\paragraph{Distillation objective} While the architecture of PISCO is similar to existing approaches, its training principle is fundamentally different. The motivation for using a distillation approach stems from an invariance principle: language models should give the same answers whether their input is compressed or not. To achieve this, we propose to use Sequence-level Knowledge Distillation (SKD) \cite{kim2016sequence}: generating labels with a teacher model rather than token-level distillation based on existing labels as done in previous works.

Specifically, given a query $q$, let $a_1,\ldots,a_r$ represent the tokens generated by the teacher based on the documents and query and $\mathbf{a_{<i}}=(a_1,\ldots,a_{i-1})$: 
\[
    a_i \sim \mathcal T(\cdot \mid d_1, \ldots, d_k,q, \mathbf{a_{<i}}).
\]
The training objective on the parameters $\theta_c$ and $\theta_d$ is the cross-entropy loss computed on the decoder conditioned on the compressed documents and the query:
\begin{align*}
        \mathbf{c_i} &= (c_i^s)_{s=1,\ldots,l} = \mathcal F_{\theta_c} (d_i,m_1,\ldots,m_l) \\
    \mathcal{L}(\theta_c, \theta_d) &= - \sum_{i=1}^{r} \log \mathcal F_{\theta_d}(a_i \mid q, \mathbf{c_1}, \ldots, \mathbf{c_k}, \mathbf{a_{<i}})
\end{align*}

Further details on this process are provided in Appendix \ref{appendix:implementation_details}. Note that the teacher-generated labels can be precomputed and re-used across different training runs.

Note that in xRAG \cite{cheng2024xrag}, the authors minimize the Kullback-Leibler (KL) divergence between the logits of the teacher and student models, with both models being teacher-forced on a reference answer. Similarly, CEPED \cite{yen2024long} minimizes a weighted combination of the KL divergence between the teacher logits  and student models and the standard cross-entropy loss on reference labels. Both methods implement in fact token-level distillation and rely on labeled data. However, token-level knowledge distillation is often less efficient than sequence-level knowledge distillation \cite{kim2016sequence} and less convenient as it requires labeled data, which is not the case for SKD.

\section{Experiments}
\label{section:experiments}
Our experiments aim to measure the performance of PISCO models \S \ref{subsection:main_results}, then to analyse the importance of training data, distillation and pretraining \S \ref{subsection:pisco_is_optimal}. Furthermore, we evaluate PISCO models generalization \S \ref{subsection:out_of_domain} to out-of-domain, multilingual data and large number of documents. Finally, we investigate how information is stored within the document embeddings \S \ref{subsection:embeddings}.

\begin{table*}[!tb]
  \centering
  \resizebox{\textwidth}{!}{ % This scales the table to the width of the page
    \begin{tabular}{lccccccc} % Added an extra column for accolade on the right
        \toprule
        & \textbf{Compression rate} & \textbf{ASQA} & \textbf{HotpotQA} & \textbf{NQ} & \textbf{TriviaQA} & \textbf{POPQA} & \textbf{Average} \\
        \midrule
        \multicolumn{8}{c}{\textbf{Decoders with no compression}} \\
        \midrule
        Mistral-7B & - & 0.74 & 0.51 & 0.69 & 0.92 & 0.70 & 0.71\\
        Llama-3.1-8B & - & 0.71 & 0.50 & 0.65 & 0.90 & 0.68 & 0.69 \\
        Solar-10.7B & - & 0.75 & 0.55 & 0.71 & 0.93 & 0.71 & 0.73 \\
        \midrule
        \multicolumn{8}{c}{\textbf{Compression Models}} \\
        \midrule
        xRAG-mistral-7B$^\dagger$ \cite{cheng2024xrag} & x128 & 0.34 & 0.27 & 0.32 & 0.77 & 0.33 & 0.40 \\
        AutoCompressor$^{\dagger\ddagger}$ \cite{chevalier2023adapting} & x4 & 0.57 & 0.31 & 0.35 & 0.70 & 0.24 & 0.43 \\
        ICAE$^\ddagger$ \cite{ge2023context} & x4 & 0.47 & 0.29 & 0.42 & 0.78 & 0.43 & 0.48 \\
        DODO$^\ddagger$ \cite{qin2024dodo}  & x10 & 0.52 & 0.38 & 0.48 & 0.82 & 0.47 &  0.53   \\
        COCOM \cite{rau2024context} & x16 & 0.63 & 0.46 & 0.58 & 0.89 & 0.48 & 0.61 \\
        PISCO - Mistral$^\Delta$ (Ours) & x16 & 0.72 & 0.48 & 0.65 & 0.90 & 0.66 & 0.68 \\
       % PISCO - Mistral \textit{(+ pre-training)} & x16 & 0.71 & 0.48 & 0.64 & 0.90 & 0.66 & 0.68 \\
        PISCO - Mistral (x128) (Ours) & x128 & 0.68 & 0.46 & 0.61 & 0.89 & 0.55 & 0.64 \\
        PISCO - Llama (Ours) & x16 & 0.72 & 0.50 & 0.64 & 0.91 & 0.66 & 0.69 \\
        PISCO - Solar (Ours) & x16 & \textbf{0.78} & \textbf{0.57} & \textbf{0.70} & \textbf{0.94} & \textbf{0.71} & \textbf{0.74}\\
        \bottomrule
    \end{tabular}
    }
    \caption{\textbf{Performance (accuracy) on general domain QA with 5 retrieved documents.} PISCO x16 models, built using Mistral, Llama, and Solar decoders, \textbf{outperform all other compression models across all datasets and closely match the performance of their uncompressed counterparts}. $^\dagger$ methods limited to one context. $^\ddagger$ methods using Llama-2-8b. $^\Delta$ achieves a x5.7 inference speed up compared to Mistral-7B.}
    \label{table:main_results}
\end{table*}

\subsection{Experimental details} 
We run experiments using Mistral-7B-instruct \cite{jiang2023mistral}\footnote{\href{https://huggingface.co/mistralai/Mistral-7B-Instruct-v0.2}{huggingface/mistralai/Mistral-7B-Instruct-v0.2}}, LLama-3.1-8B-instruct \footnote{\href{https://huggingface.co/meta-llama/Llama-3.1-8B-Instruct}{huggingface/meta-llama/Llama-3.1-8B-Instruct}} and SOLAR-10.7B-Instruct \footnote{\href{https://huggingface.co/upstage/SOLAR-10.7B-Instruct-v1.0}{huggingface/upstage/SOLAR-10.7B-Instruct-v1.0}} as different backbones for PISCO. Our training set of questions \footnote{\href{https://huggingface.co/datasets/dmrau/multi_qa}{huggingface/datasets/dmrau/multi\_qa}} is taken from \cite{rau2024context}: it consists of 453k questions based on documents from Wikipedia-KILT \cite{petroni2020kilt} which we preprocess in chunks of 128 tokens and use as our database collection \footnote{\href{https://huggingface.co/datasets/dmrau/kilt-128}{huggingface/datasets/dmrau/kilt-128}}. For each question, we search the first top-k documents and feed them to a teacher LLM to obtain the silver label used for distillation. During training, the number of retrieved documents k is set to 5. Each document is compressed into $l$ embedding vectors where $l$ is fixed for each PISCO model. PISCO models with compression rate 16 use $8$ memory embeddings per document.

All experiment details, including the choice of the retriever, reranker and prompts are provided in Appendix \ref{appendix:prompt} and Appendix \ref{appendix:training_hyperparameters}. Trainings and evaluations were performed using the Bergen \cite{rau2024bergen} library.

\subsection{Main results}
\label{subsection:main_results}

After training, we first evaluate the PISCO models on general knowledge QA tasks: Natural Questions \cite{kwiatkowski2019natural}, TriviaQA \cite{joshi2017triviaqa}, HotpotQA \cite{yang2018hotpotqa}, ASQA \cite{stelmakh2022asqa}, and PopQA \cite{mallen2022not} datasets. Our main evaluation metric is the accuracy --also called match in the QA context-- which we define as 1 if the normalized label is found within the normalized prediction and 0 otherwise, as detailed in Appendix \ref{appendix:normalize}.

Results are shown on Table \ref{table:main_results}. PISCO largely outperforms the other compression methods with +8\% accuracy on average compared to COCOM with Mistral backbone. PISCO - Mistral trained with 128 compression rate outperforms xRAG by more than $20\%$. \textbf{In fact, all PISCO models with compression rate 16 are very close (0-3\%) to their uncompressed backbones.} Most notably, PISCO-Solar outperforms Solar, showing that the compression can have a de-noising effect, discarding irrelevant information.

By changing the number of memory tokens $l$ prompted to the compressor $\mathcal F_{\theta_c}$, we can train models with different compression rates. Figure \ref{fig:intro_figure} shows the average accuracy for PISCO-Mistral models with compression rates between x2 and x128. Performance decreases gradually as the compression rate increases up to 16, after which the decline becomes more pronounced.

\begin{figure}[!tb]
    \centering
    \includegraphics[width=\columnwidth]{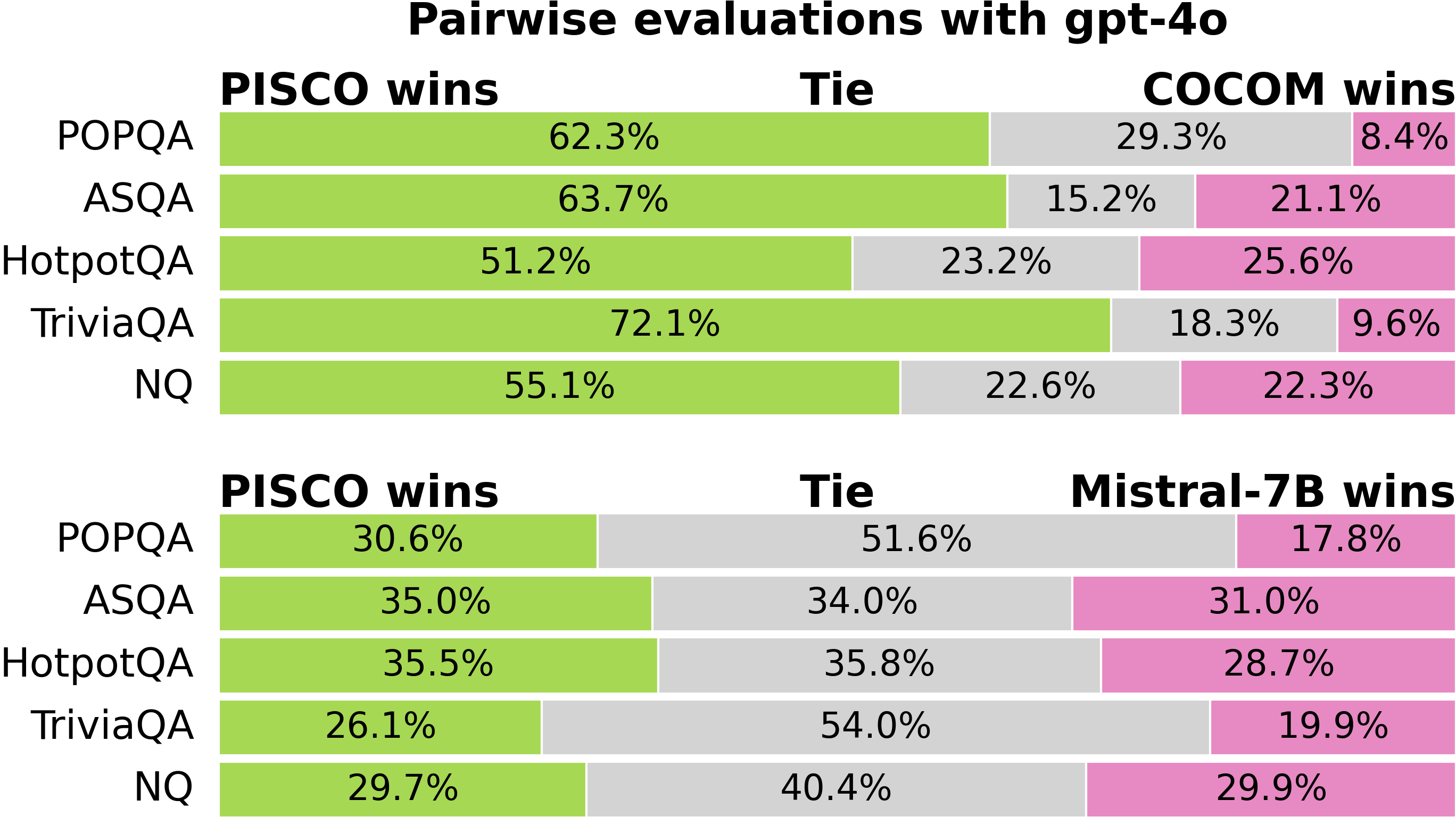} 
    \caption{Pairwise comparison with GPT-4o shows that PISCO, utilizing the Mistral-7B backbone, outperforms COCOM across all datasets. It performs comparably to Mistral-7B while achieving a 16x compression rate.}
    \label{fig:pairwise_gpt}
\end{figure}

Then, we use gpt-4o \cite{hurst2024gpt} and its strong ability to assess semantic content to perform pairwise comparisons \cite{dubois2024length} between generated answers to compare models. Results are shown on Figure \ref{fig:pairwise_gpt}. It shows that PISCO-Mistral outperforms COCOM model and is on par with its uncompressed backbone. The exact setup for this evaluation is detailed in appendix \ref{appendix:gpt_evaluation}.

\paragraph{Computational efficiency} For Mistral-7B and PISCO-Mistral, we measure FLOPS, maximum batch size and inference time on an A100 gpu. We perform these measures on a 128-token query and 5 128-token documents, forcing the generation of a 32-token answer. Results are shown on Table \ref{table:efficiency_mistral}. PISCO-Mistral is 4 times faster than Mistral-7B, it supports a maximum batch size four times larger. PISCO-Mistral with a compression rate of 128 is 5 times faster than Mistral-7B for generation.

\begin{table}[!tb]
     \centering
     \setlength{\tabcolsep}{3pt} % Reduce horizontal padding (default is 6pt)
    \begin{tabular}{lrrr}
    \toprule
    Model & GFlops &  Time(s) & Max batch size \\
    \hline
    \hline
    Mistral-7B & 11231  & 0.26 & 256 \\
    PISCO-Mistral & 2803 & 0.06 & 1024 \\
    PISCO-Mistral (x128) & 2312 & 0.05 & >1024 \\
    \bottomrule
    \end{tabular}
    \caption{Efficiency of Mistral-7B and PISCO-Mistral (x16 and x128).}
    \label{table:efficiency_mistral}
\end{table}

\subsection{Analysis of Training Data and Tasks for Compression}
In this section, we aim to give evidences justifying the design choices of the PISCO approach.

\label{subsection:pisco_is_optimal}

%\subsection{Pretraining has little benefits}
%\label{subsection:pretraining}
\paragraph{Pretraining has little benefits.}
A potential way to improve compression models was by improving pretraining with new or refined pretraining tasks. To explore this, we conducted experiments using pretraining on 10B tokens extracted from FineWeb\footnote{\href{https://huggingface.co/datasets/HuggingFaceFW/fineweb}{huggingface./datasets/HuggingFaceFW/fineweb}} with a variety of tasks. These tasks included auto-encoding \cite{ge2023context,rau2024context}, text continuation from compressed representations \cite{rau2024context,cheng2024xrag}, and a novel task of text continuation from a sequence of keywords within a compressed segment—enabling access to information embedded in the learned representations, aimed at mitigating a potential “lost-in-the-middle” effect. Additionally, we tested continuation from multiple documents, where the model was prompted to continue text either from within or following a designated document among several compressed ones (see Appendix \ref{appendix:pretraining_tasks}). Figure \ref{fig:ae_tc_correlation_with_rag} illustrates that, across all preliminary experiments, there is \textbf{a weak correlation between success in pretraining tasks and performance in QA}. Notably, training on auto-encoding often achieves near-perfect Rouge-L scores (>0.99) without any significant improvement in QA performance. 

\begin{figure}[!tb]
    \centering
    \includegraphics[width=\columnwidth]{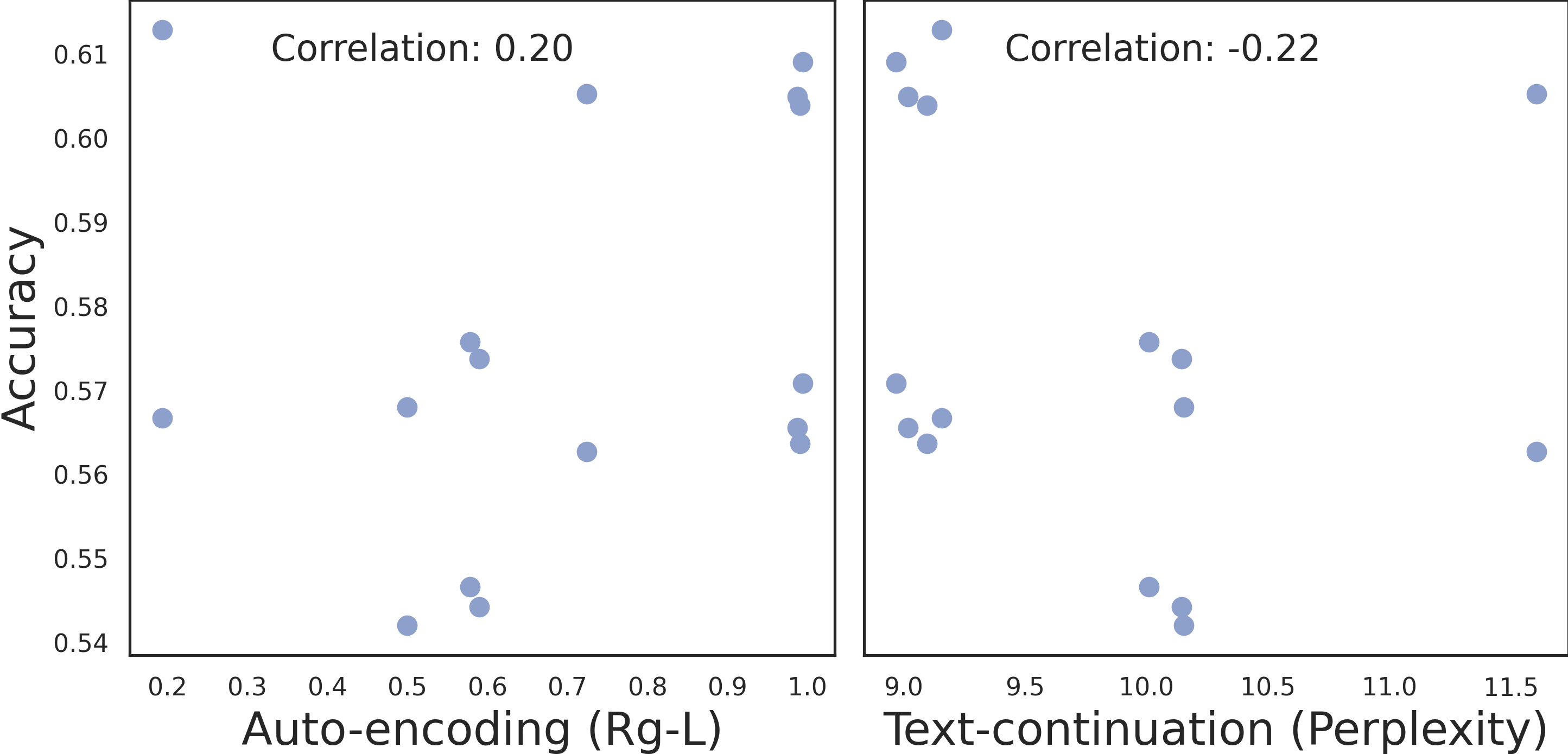} 
    \caption{Performances on pretraining tasks versus performance on RAG-QA. Correlations are very small, indicating that pretraining has only little benefits on the downstream QA task.}
    \label{fig:ae_tc_correlation_with_rag}
\end{figure}

\begin{figure}[!tb]
    \centering
    \includegraphics[width=\columnwidth]{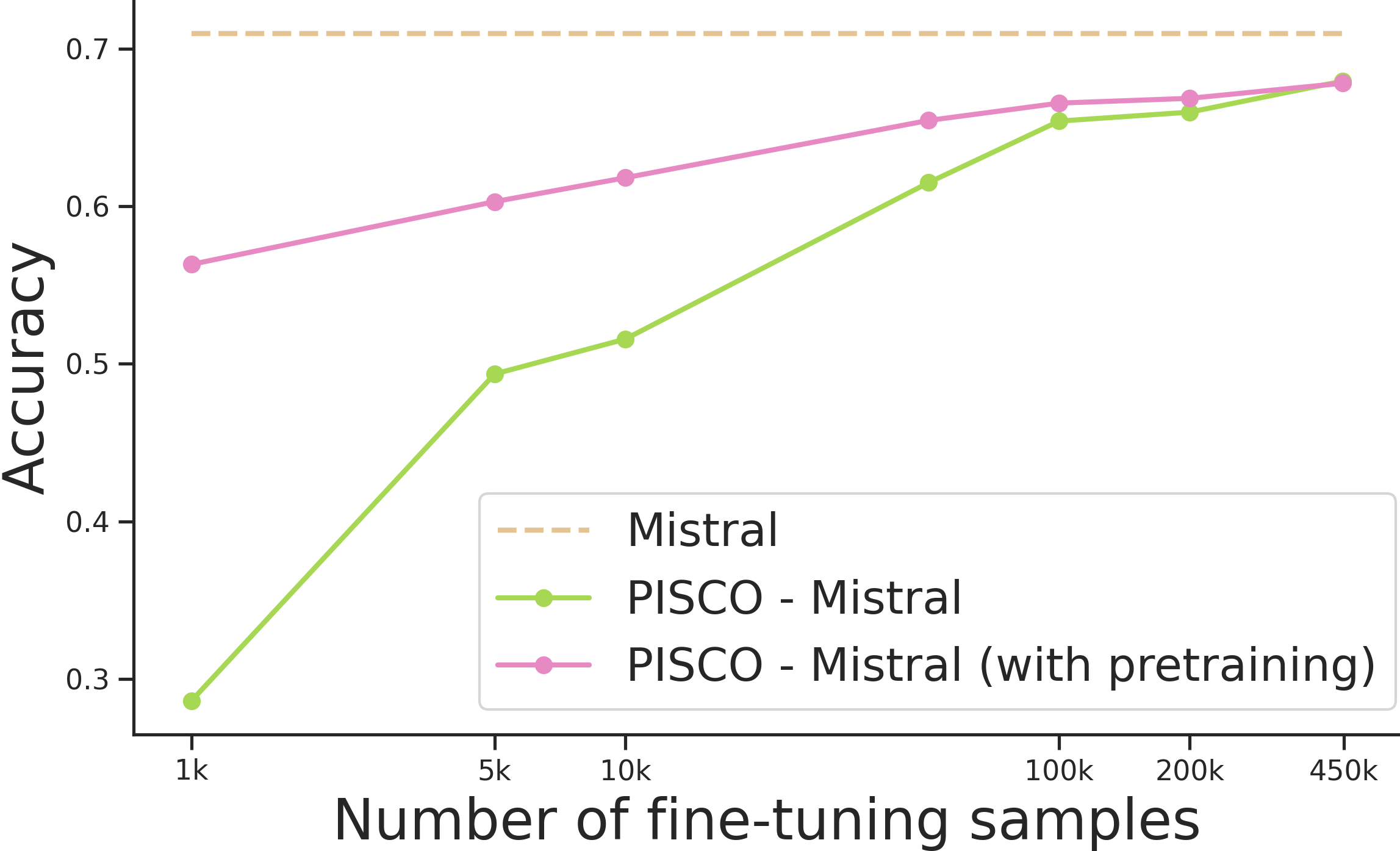} 
    \caption{Impact of the number of fine-tuning samples on performance, with and without pretraining. Pretraining only improves QA performance for low fine-tuning sample size.}
    \label{fig:variable_train_sample_with_pretraining}
\end{figure}

To analyze in detail the impact of the adopted fine-tuning strategy, we ran experiments with variable number of fine-tuning samples. We compare performances when fine-tuning is applied to a pretrained model or from scratch. Results are shown on Figure \ref{fig:variable_train_sample_with_pretraining}. Pretraining benefits final performances for low fine-tuning sample size, but it is not useful at 450k samples.

\begin{figure}[!tb]
    \centering
    \includegraphics[width=0.8\columnwidth]{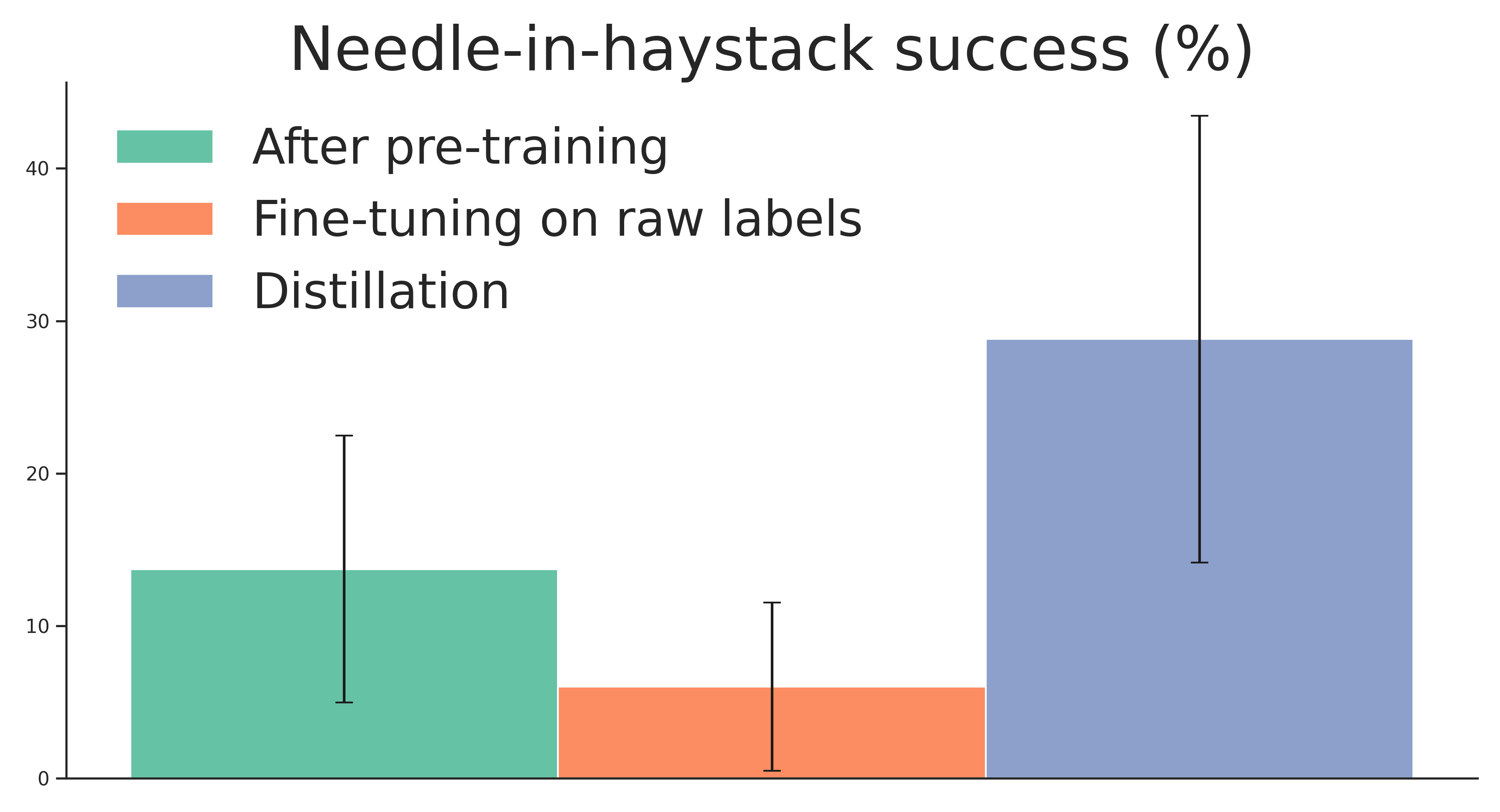} 
    \caption{Needle-in-a-haystack results from preliminary experiments: fine-tuning on raw labels hinders performance, while sentence-level distillation enhances it.}
    \label{fig:nih}
\end{figure}

\paragraph{Needle-in-a-Haystack analysis.}
Secondly, we analyzed the different model behaviors on a needle-in-a-haystack test \cite{nih_github}, an accessible proxy for RAG-QA that effectively measures the model’s retrieval and localization capabilities, crucial for accurate question-answering on large datasets. Interestingly, while pretraining enables some success on the needle-in-a-haystack task, fine-tuning on the raw labels used in \cite{rau2024context, cheng2024xrag} diminishes this capability, as illustrated in Figure \ref{fig:nih}. This result highlighted the need for a better fine-tuning approach. Higher-quality labels from a teacher LLM emerged as a promising solution: providing more informative signals during fine-tuning \cite{ho2022large,ren2024learn}. Early experiments shown on Figure \ref{fig:nih} and main results shown on Table \ref{table:main_results} indeed confirm the benefits.

\begin{figure}[!tb]
    \centering
    \includegraphics[width=\columnwidth]{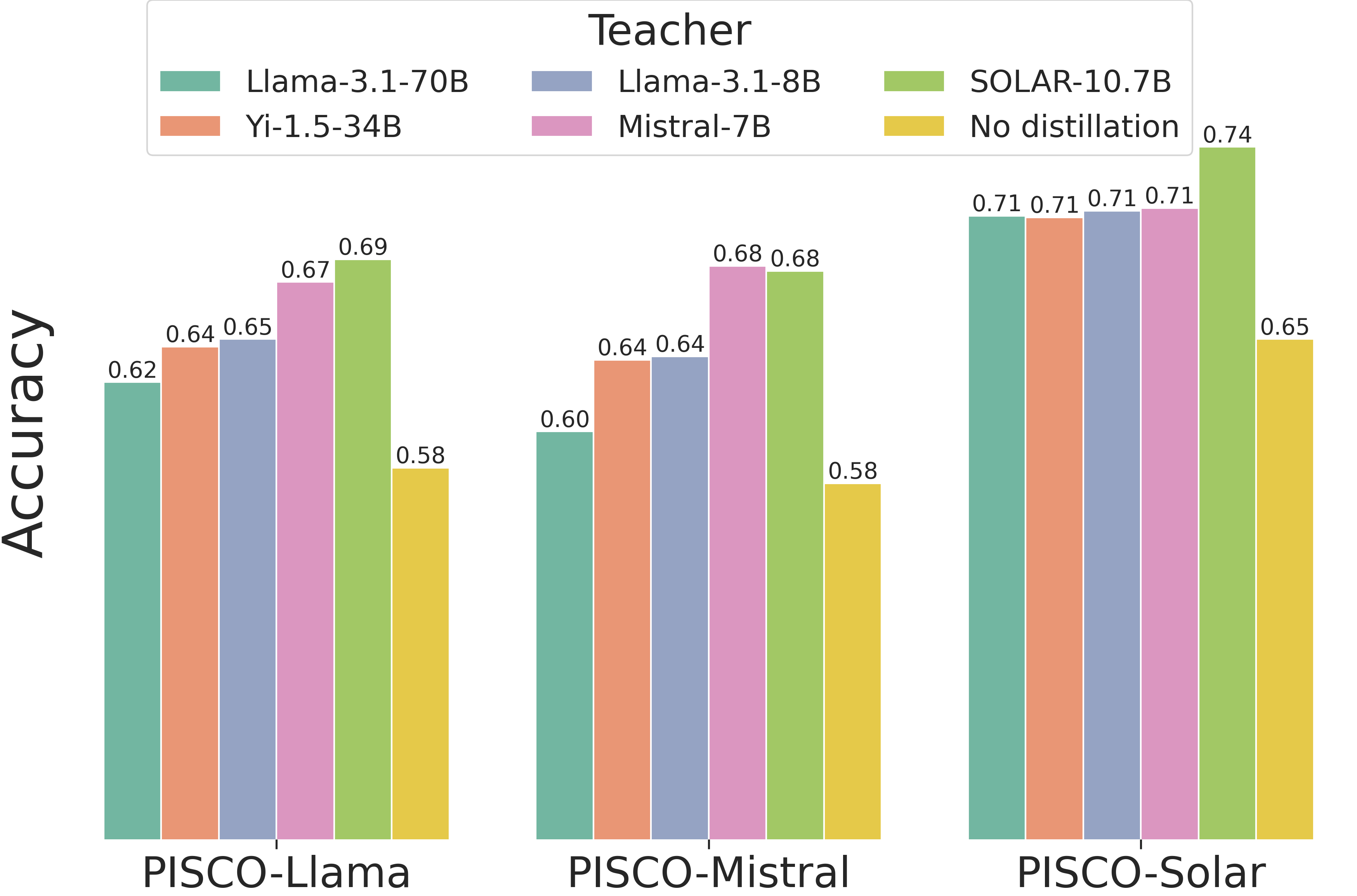} 
    \caption{Impact of the choice of teacher on PISCO. No distillation means supervised fine-tuning on the existing labels as in \cite{rau2024bergen}.}
    \label{fig:teacher_impact}
\end{figure}

%\subsection{Impact of Teacher and Labels Quality}
\paragraph{Impact of Teacher and Labels Quality.}
To understand the impact of the teacher LLM we train PISCO models with varying teachers and without distillation. Average accuracy in general domain QA for each obtained model are shown on Figure \ref{fig:teacher_impact}. Interestingly, the best-performing teachers are generally Mistral-7B or Solar-10B models, not necessarily the stronger teachers, as found in \cite{xu2024stronger}. A manual analysis of the labels for each teacher suggests that these models often include justifications for their answers based on the given contexts: training the PISCO models to replicate this reasoning process may account for the observed performance improvements, as shown in \cite{ho2022large, ren2024learn}. Note that PISCO-Solar models are very robust across all teachers. Removing the sequence-level distillation (i.e. doing supervised fine-tuning on the existing labels from \cite{rau2024context}) leads to much lower results.

To summarize our analysis, we have shown that there is little transfer between pretraining tasks and downstream RAG-QA performance, thus justifying why pretraining is not key for compression models. Secondly, we have seen that standard fine-tuning on raw labels is problematic since it leads to poor results in needle-in-a-haystack test. Then, we showed that SKD with a teacher LLM solves this problem. Finally, we tested various teacher LLMs, recovering some of the findings of \cite{ho2022large, xu2024stronger, ren2024learn}. Overall, these analysis explain the rationale which led to each design decisions of the training pipeline for PISCO.

\subsection{Generalization Evaluation}
\label{subsection:out_of_domain}
\paragraph{Increasing the number of documents.} 

\begin{figure}[!tb]
    \centering
    \includegraphics[width=\columnwidth]{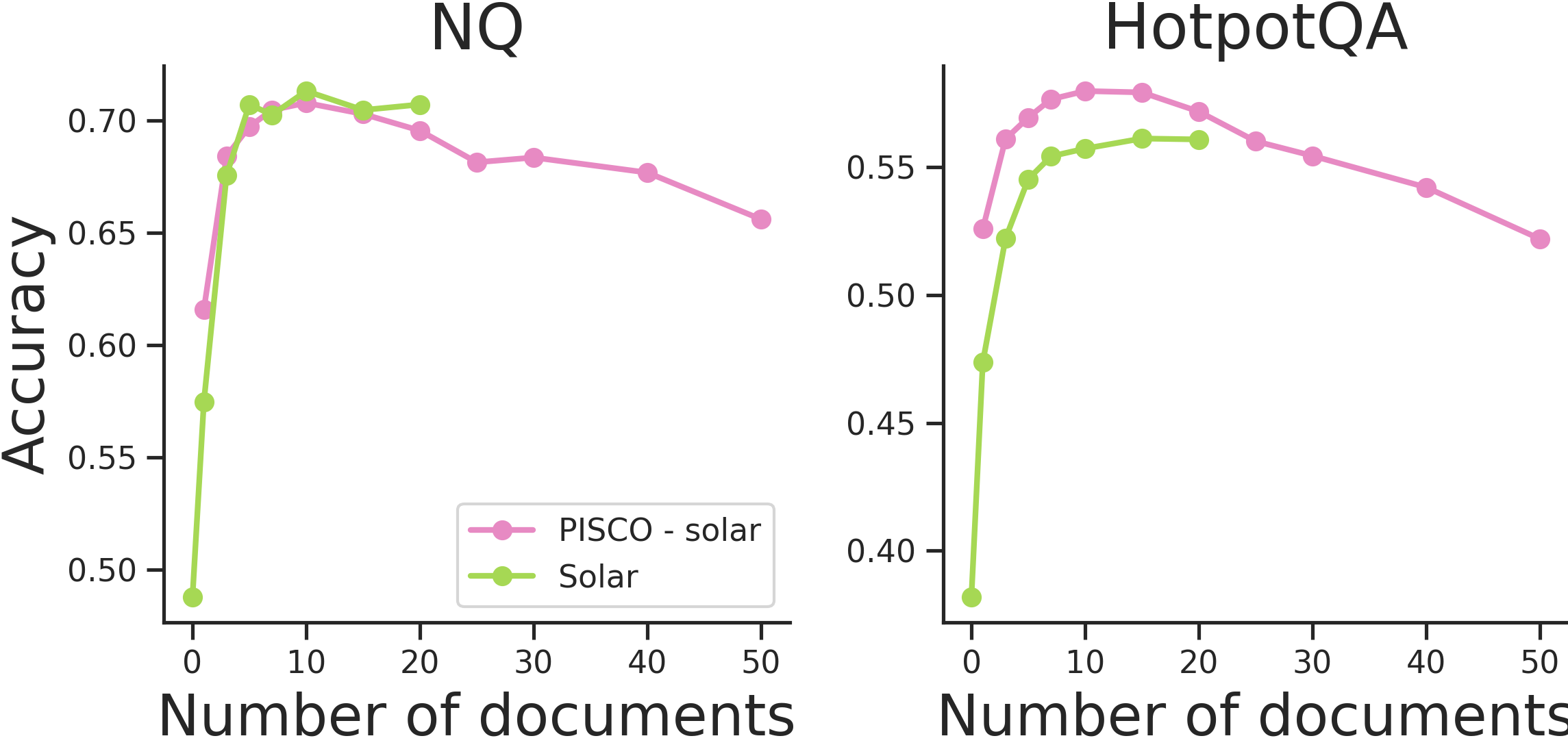} 
    \caption{Accuracy of PISCO-Solar compared to Solar with increasing number of documents.}
    \label{fig:large_top_k}
\end{figure}

To evaluate the PISCO models’ ability to handle a large volume of documents, we conduct inference on PISCO - Solar with document sets ranging from 1 to 50 on NQ and HotpotQA. Results are presented in Figure \ref{fig:large_top_k}. For comparison, we include the base SOLAR model, which can process only up to about 20 documents due to its 4096-token context length limit. PISCO’s performance aligns closely with the base model, with a gradual decrease beyond 20 documents, a common trend in RAG models \cite{jin2024long}, which could be addressed by increasing $k$ during fine-tuning (all of our experiments train with $k=5$ documents).

\paragraph{Out-of-domain.}
We evaluate whether PISCO models generalize to unseen domains using a diverse set of datasets (details in Table \ref{table:out_of_domain_datasets} in the appendix). The results are shown in Table \ref{table:out_of_domain_results}. Results in Table \ref{table:out_of_domain_results} show that PISCO models generalize nearly as effectively as their backbone decoders, with some performance drops (-3–10\%) on the ParaphraseRC task. This indicates robust compression capabilities, showing PISCO models do not rely on memorizing the general knowledge in the KILT collection.

\begin{table*}
  \centering
  \resizebox{\textwidth}{!}{ % This scales the table to the width of the page
    \begin{tabular}{lcccccccc|ccc}
    %\toprule
    %& & & & \multicolumn{5}{c}{RobustQA} & \multicolumn{5}{c}{Multilingual}\\
    & & & & \multicolumn{5}{c}{$\overbrace{\hspace{19em}}^{\text{\large\textbf{RobustQA}}}$} & 
    \multicolumn{3}{c}{$\overbrace{\hspace{8em}}^{\text{\large\shortstack{\textbf{Multilingual QA}}}}$} \\
    
    %\cline{5-9}\cline{10-12}
    Dataset & Bio-QA  & Covid & ParaphraseRC & Lifestyle & Writing & Science & Recreation & Tech & FR & KO & RU\\
    Metric & Recall & F1 & Accuracy & F1 & F1 & F1 & F1 & F1 & \multicolumn{3}{c}{recall-3gram} \\
    \midrule
    \midrule
    Llama & 0.26 & \textbf{0.17} & 0.48 & 0.25 & 0.23 & 0.25 & 0.25 & 0.23 & 0.56 & 0.28 & 0.47 \\
    PISCO - Llama & 0.24 & 0.12 & 0.38 & 0.28 & 0.27 & \textbf{0.26} & \textbf{0.26} & \textbf{0.26} & 0.54 & 0.24 & 0.44 \\
    \midrule
    Mistral & 0.27 & 0.13 & 0.49 & 0.28 & 0.27 & \textbf{0.26} & 0.25 & 0.25 & 0.57 & 0.26 & 0.40 \\
    PISCO - Mistral & 0.26 & 0.11 & 0.38 & 0.28 & 0.27 & \textbf{0.26} & 0.25 & \textbf{0.26} & 0.52 & 0.17 & 0.35 \\
    \midrule
    Solar & 0.28 & 0.14 & \textbf{0.50} & 0.28 & 0.27 & \textbf{0.26} & \textbf{0.26} & 0.25 & 0.59 & 0.20 & 0.52 \\
    PISCO - Solar & \textbf{0.29} & 0.10 & 0.47 & \textbf{0.29} & \textbf{0.28} & 0.25 & 0.25 & \textbf{0.26} & 0.60 & 0.16 & 0.48 \\
    \bottomrule
    \end{tabular}
    }
    \caption{Out-of-domain and multilingual QA performance. PISCO models generalize well to new domains and languages, achieving performance comparable to their uncompressed decoders. See Table \ref{table:out_of_domain_datasets} for datasets details.}
    \label{table:out_of_domain_results}
\end{table*}

%\subsection{Multilingual evaluation}
\paragraph{Multilinguality.}
We evaluate whether PISCO models generalize to unseen languages using the MKQA dataset \cite{longpre2021mkqa}. The experiments use the bge-m3 retriever and the recall-3gram metric, more resilient to language variation \cite{chirkova2024zero}. We choose a latin language (french) as well as Korean and Russian. Note that the PISCO backbones Llama and Mistral are not strong multilingual models, but these experiments serve mostly to analyze the compression behavior. Results are shown on Table \ref{table:out_of_domain_results} (right). PISCO models seem to generalize fairly well to other languages, with still a small drop compared to their backbones. Further analysis is needed to determine whether the drop is due to compression or language generation limitations.

%\begin{table}
%  \centering
%  \resizebox{0.3\textwidth}{!}{ % This scales the table to the width of the page
%    \begin{tabular}{lrrrrrr}
%    \toprule
%     Language & FR & KO & RU\\
%    \midrule
%    \midrule
%   Llama & 0.56 & 0.28 & 0.47 \\
%    PISCO - Llama & 0.54 & 0.24 & 0.44 \\
%    \midrule
%    Mistral & 0.57 & 0.26 & 0.40 \\
%    PISCO - Mistral & 0.52 & 0.17 & 0.35 \\
%    \midrule
%    Solar & 0.59 & 0.20 & 0.52 \\
%    PISCO - Solar & 0.60 & 0.16 & 0.48 \\
%    \bottomrule
%    \end{tabular}
%    }
%    \caption{Multilingual performances (recall-3gram) on MKQA.}
%    \label{table:multilingual_results}
%\end{table}

%\begin{table}
%  \centering
%  \resizebox{0.5\textwidth}{!}{ % This scales the table to the width of the page
%    \begin{tabular}{lrrrrrr}
%    \toprule
%     Query Language & FR & KO & RU & FR & KO & RU\\
%     Retrieval language & \multicolumn{3}{c}{English} |& \multicolumn{3}{c}{User language}\\
%    \midrule
%    \midrule
%    Llama & 0.61 & 0.26 & 0.52 & 0.56 & 0.28 & 0.47 \\
%    PISCO - Llama & 0.50 & 0.14 & 0.20 & 0.54 & 0.24 & 0.44 \\
%    \midrule
%    Mistral & 0.62 & 0.19 & 0.31 & 0.57 & 0.26 & 0.40 \\
%    PISCO - Mistral & 0.46 & 0.07 & 0.08 & 0.52 & 0.17 & 0.35 \\
%    \midrule
%    Solar & 0.66 & 0.19 & 0.58 & 0.59 & 0.20 & 0.52 \\
%    PISCO - Solar & 0.54 & 0.09 & 0.25 & 0.60 & 0.16 & 0.48 \\
%    \bottomrule
%    \end{tabular}
%    }
%    \caption{Multilingual performances (recall-3gram) on MKQA.}
%    \label{table:multilingual_results}
%\end{table}

To sum-up, we showed that \textbf{PISCO models are robust to the number of documents} used, \textbf{generalize well to unseen domains} beyond the Wikipedia collections, and that \textbf{compression works relatively well for unseen languages}. Overall, it shows that PISCO models are strong language models for RAG.

\subsection{Document embeddings analysis}
\label{subsection:embeddings}

\begin{figure}[!tb]
    \centering
    \includegraphics[width=0.8\columnwidth]{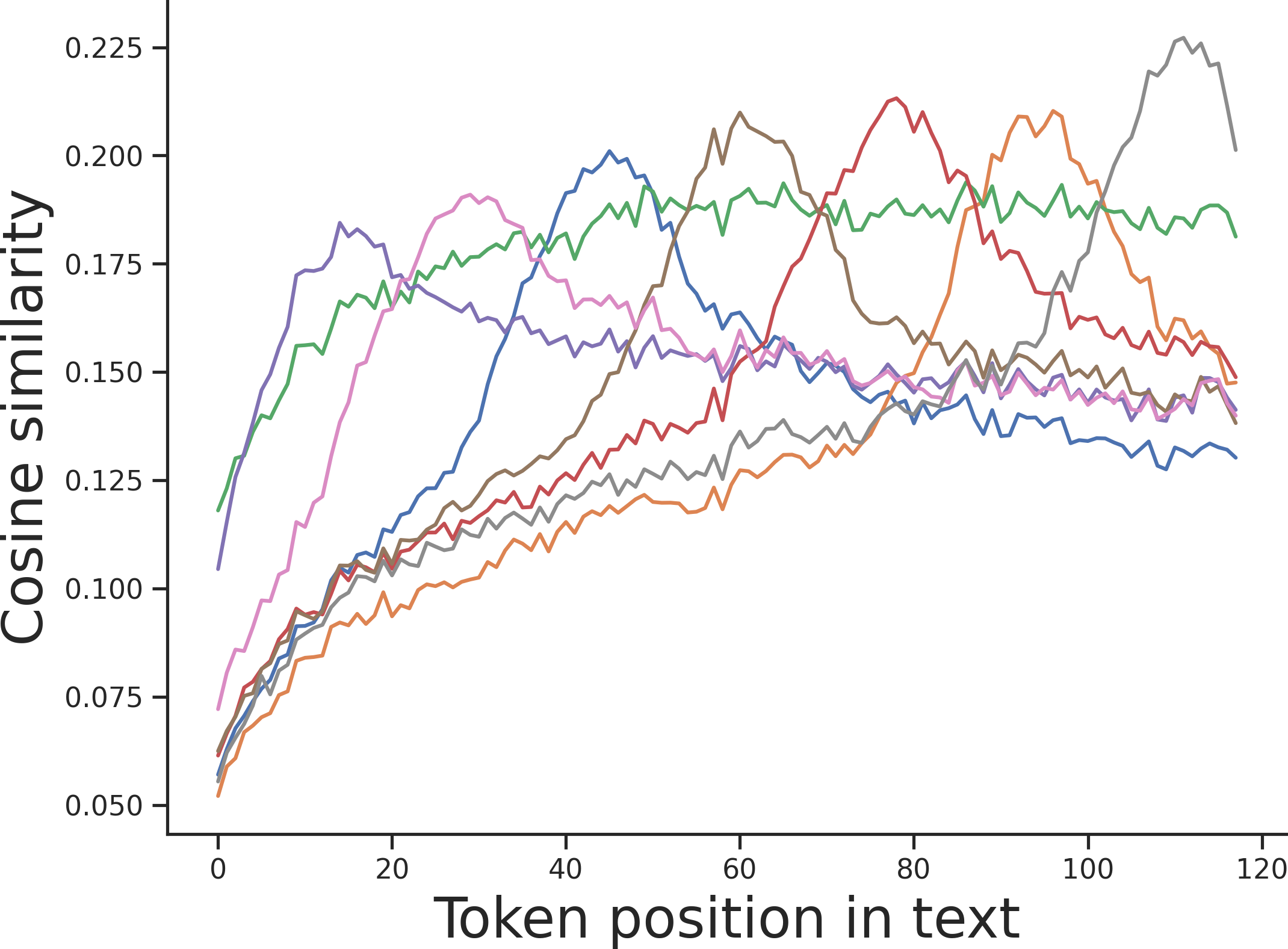} 
    \caption{Average cosine similarities between each of the $l=8$ document embeddings and individual document tokens reveals spatial specialization in the memory embeddings, with each focusing on neighboring positions in the encoded document.}
    \label{fig:cosine_similary_embeddings}
\end{figure}

\begin{figure}[!b]
    \begin{tcolorbox}[colback=gray!5!white,
    colframe=gray!75!black,
    fontupper=\small,
    ]
    Pisco is a \textcolor{algocolor2}{colorless} or \textcolor{algocolor1}{yellowish-to-amber-colored} \textcolor{algocolor2}{spirit} \textcolor{algocolor2}{produced} \textcolor{algocolor4}{in} \textcolor{algocolor4}{winemaking} regions of Peru \textcolor{algocolor0}{and} Chile. Made by \textcolor{algocolor6}{distilling} \textcolor{algocolor6}{fermented} grape \textcolor{algocolor6}{juice} \textcolor{algocolor1}{into} a \textcolor{algocolor4}{high-proof} \textcolor{algocolor2}{spirit,} it was \textcolor{algocolor0}{developed} by \textcolor{algocolor4}{16th-century} \textcolor{algocolor0}{Spanish} \textcolor{algocolor0}{settlers} as \textcolor{algocolor0}{an} \textcolor{algocolor0}{alternative} \textcolor{algocolor1}{to} orujo, a \textcolor{algocolor5}{pomace} \textcolor{algocolor0}{brandy} that was being \textcolor{algocolor5}{imported} from Spain. It had the \textcolor{algocolor0}{advantages} of being \textcolor{algocolor2}{produced} from \textcolor{algocolor0}{abundant} \textcolor{algocolor3}{domestically} grown \textcolor{algocolor3}{fruit} \textcolor{algocolor0}{and} \textcolor{algocolor3}{reducing} the \textcolor{algocolor1}{volume} of \textcolor{algocolor1}{alcoholic} \textcolor{algocolor7}{beverages} \textcolor{algocolor0}{transported} \textcolor{algocolor1}{to} \textcolor{algocolor1}{remote} \textcolor{algocolor7}{locations.}
    \end{tcolorbox}
    \caption{Each colored word in this text appears in the top-10 token attributed with logit lens in the document embeddings. Most words appear.}
    \label{fig:text_example}
\end{figure}

To better understand how compression works, we compute, on a set of documents, the cosine similarity between the $l$ embeddings and each document token. Interestingly, Figure \ref{fig:cosine_similary_embeddings} shows there is a spatial specialization of the embeddings, each attending preferably to some part of the text. This specialization does not occur when the decoder is kept frozen. Then, we used the logits lens \cite{nostalgebraist2020logit} to determine, for each document embedding, the top tokens when applying the LLM head. Figure \ref{fig:text_example} shows, for some text, how often text tokens correspond to one of these attributed tokens. Additional results are shown on Table \ref{table:logits_attribution}.

Second, we gather all embeddings of all documents as well as token embeddings and visualize using t-SNE, in the input space of the decoder $\mathcal F_{\theta_d}$. Result is shown in the appendix on Figure \ref{fig:tnse_embedding}. The document embeddings tend to lie outside of the document tokens embeddings. This underscores the need of fine-tuning the decoder to exploit newly formed compressed representations: an effective pipeline leverages new areas of the embedding space, with different semantic content.

\section{Conclusion}
\label{section:discussion}
We proposed PISCO, the first compression method for RAG which enables large compression rates with little to no accuracy loss. Our analysis and ablations revealed the ineffective transfer between pretraining and the question answering task for compression models. We also showed the importance of training labels for compression: using an appropriate teacher LLM for distillation is key. Given the strong evidences of robustness and accuracy, PISCO models may be used as drop-in replacement for existing RAG systems currently relying on their uncompressed backbones. Adopting PISCO would reduce inference costs and latency with minimal performance impact.

%\newpage

%\section{Limitations}

%While achieving state-of-the-art performances for context compression for question-answering with retrieval augmented generation, our work has some limitations.

%First, we only tested the compression method in a QA setting: it is clear that the compressed embeddings carry a lot of semantic value and there are reasons to believe it should enable to deal with more complex tasks such as summarization, long chats, long contexts via compression etc and this would form a natural continuation of our work. 

%Second, at variance with \cite{rau2024context}, we do not propose in this contribution a light version of the compressor to be used in an online setting (i.e. compressing documents on the fly with a compression method sufficiently fast to accelerate the full RAG pipeline overall). Therefore our approach is only valuable when the collection of documents can be compressed once beforehand and one expects a sufficiently large volume of queries to compensate for the cost of the compression (NB: our tests show that compressing 1 million documents takes approximately 3h on a high-end GPU).

%Third, our experiments suggest some multilingual generalization of the model abilities but to enable strong performances, we should augment our training set with multilingual data.

{
    \small
    \bibliographystyle{ieeenat_fullname}
    \bibliography{biblio}
}

\clearpage
\appendix

\section{Implementation details}
\label{appendix:implementation_details}

Both for teacher-label generation and student evaluation, generation is done using greedy decoding, limited to a maximum of $128$ tokens. Both for training and evaluation, documents are retrieved using Splade-v3 \cite{lassance2024splade} and Debertav3 \cite{he2021debertav3}. They are prompted from most relevant to less relevant according to the reranking scores.

\section{Training Hyper-parameters}
\label{appendix:training_hyperparameters}

Table \ref{table:hyperparams} gives the hyper-parameters we used for training. Note that on top of the LoRA adapters, the embeddings of the memory tokens given as input to the encoder (as in \cite{rau2024context, chevalier2023adapting}) are also optimized, adding an extra $l \times$ model\_hidden\_size trainable parameters to each PISCO model (much less than the number of parameters in the adapters).

\begin{table}[!tb]
    \centering
    \begin{tabular}{|ll|}
    \hline
    \textbf{Hyperparameter} & \textbf{Value} \\ \hline
    Batch Size & 128 \\
    LR & $1 \times 10^{-4}$ \\
    LR scheduler & linear \\
    optimizer & AdamW \\
    Epochs & 1 \\
    Max Tokens Teacher Generation & 128 \\ 
    LoRA Layers ($r$) & all-linear \\ 
    LoRA Rank ($r$) & 16 \\ 
    LoRA Dropout & 0.1 \\ 
    LoRA Alpha & 32 \\ 
    LoRA Rank ($r$) & 16 \\ 
    Weight Decay & 0.1 \\ 
    Warmup Ratio & 0.05 \\ 
    Max Gradient Norm & 1.0 \\ 
    Documents max tokens & 128 \\
    \hline
    \end{tabular}
    \caption{Fine-tuning Hyper-parameters.}
    \label{table:hyperparams}
\end{table}

\section{Main prompt}
\label{appendix:prompt}

\noindent Below is the prompt used in our experiments. <DOC> is replaced by the corresponding document compressed embeddings before the generation while <QUESTION> is replaced with the query $q$. It is formatted as an instruction prompt to the instruction-tuned models.

\begin{tcolorbox}[colback=gray!5!white,colframe=gray!75!black,title=Main prompt]
\textbf{system}: "You are a helpful assistant. Your task is to extract relevant information from provided documents and to answer to questions as briefly as possible."

\textbf{user}: "Background:\newline
<DOC><SEP><DOC>\ldots<SEP><DOC>
\newline
Question: <QUESTION>"
\end{tcolorbox}

\section{Effect of prompts on PISCO models}
\label{appendix:prompt_effect}

We evaluate the robustness of PISCO models to prompt variations by testing with modified versions of the prompt shown in \ref{appendix:prompt}. The results in Table \ref{table:prompt_effect} show minimal performance differences, indicating that the models are stable with different prompts and do not overfit to the specific prompt used during training. Notably, Prompt 3 provides no guidance to the model beyond the information in the documents.

\begin{table}[!tb]
 \centering
\begin{tabular}{clll}
\toprule
 & NQ & TriviaQA & HotpotQA \\
\hline
\hline
Prompt 1 & 0.697 & 0.937 & 0.569 \\
Prompt 2 & 0.694 & 0.942 & 0.571 \\
Prompt 3 & 0.684 & 0.940 & 0.567 \\
Prompt 4 & 0.694 & 0.937 & 0.572 \\
\bottomrule
\end{tabular}
\caption{Effect of 4 different prompts on match for three datasets.}
\label{table:prompt_effect}
\end{table}

\begin{tcolorbox}[colback=gray!5!white,colframe=gray!75!black,title=Other prompts]
\begin{itemize}[left=0pt]
    \item \textbf{system}: "Refer to the background document and answer the questions:"
    
    \textbf{user}: "Background:\newline
    <DOC><SEP>\ldots<DOC>
    \newline
    Question: <QUESTION>"
    \item \textbf{system}: ""
    
    \textbf{user}: "Background:\newline
    <DOC><SEP>\ldots<DOC>
    \newline
    Question: <QUESTION>"
    \item \textbf{system}: ""
    
    \textbf{user}: "<DOC><SEP>\ldots<DOC>
    Question: <QUESTION>"
\end{itemize}
\end{tcolorbox}

\section{Pairwise comparison using gpt-4o}
\label{appendix:gpt_evaluation}

To compare answers generated by different methods in a more precise way that using the accuracy metric, we use gpt-4o with the following prompt, inspired from Alpaca-eval \cite{dubois2024length}. Evaluations were run using gpt-4o-2024-11-20. To limit costs, only a 1000 samples were used for each dataset. Answer positions in the prompt were randomly switched to prevent position bias.
        
\begin{tcolorbox}[colback=gray!5!white,colframe=gray!75!black,title=Gpt pairwise comparison prompt]
\textbf{system}: "You are a helpful assistant, that ranks models by the quality of their answers. Please act as an impartial judge. Do not allow the length of the responses to influence your evaluation. Be as objective as possible."

\textbf{user}: "Here is a question, a ground truth answer, an AI-generated answer 1 and an AI-generated answer 2. Which answer is the most correct one ? Simply answer {{1}} if the first is better, {{2}} if the second is better and {{3}} if it's a tie.

Question: <QUESTION>.

Ground truth answer: <REF\_ANSWER>.

Answer 1: <ANSWER1>.

Answer 2: <ANSWER2>."

\end{tcolorbox}

\section{Out-of-domain datasets}

Table \ref{table:out_of_domain_datasets} provides details on the out-of-domain datasets used and the primary evaluation metric for each. We use the F1 score for RobustQA test suites, given the extended format of the reference answers.

\begin{table*}[!tb]
    \centering
    \begin{tabular}{lllll}
    \toprule
    \multicolumn{2}{l}{Dataset} & Document collection & Evaluation metric & N \\
    \hline \hline
     \multicolumn{2}{l}{Bio-QA \cite{krithara2023bioasq}} & Pubmed & Recall & 3837 \\ \hline
     \multicolumn{2}{l}{Covid \cite{moller2020covid}} & CORD-19 & F1 & 2019 \\ \hline
     \multicolumn{2}{l}{ParaphraseRC \cite{saha2018duorc}} & ParaphraseRC & Accuracy & 13111 \\ \hline
    \multirow{5}{*}{RobustQA \cite{han2023robustqa}} & Lifestyle 1 & \multirow{5}{*}{LoTTE} & \multirow{5}{*}{F1} & 2198 \\ \cline{2-2} \cline{5-5}
     & Writing &  &  & 2694 \\ \cline{2-2} \cline{5-5}
     & Science &  &  & 1404 \\ \cline{2-2} \cline{5-5}
     & Recreation &  &  & 2090 \\ \cline{2-2} \cline{5-5}
     & Technology &  &  & 2064 \\ \cline{2-2} \cline{5-5}
     \hline
     \multicolumn{2}{l}{MKQA \cite{longpre2021mkqa}} & Wikipedia\footnote{\href{https://huggingface.co/datasets/wikimedia/wikipedia}{https://huggingface.co/datasets/wikimedia/wikipedia}} & recall-3gram & 10000 \\
    \hline
    \end{tabular}
    \caption{Out-of-domain datasets characteristics. We use F1 for RobustQA and CovidQA since their labels are long and not suitable for computing the accuracy directly.}
    \label{table:out_of_domain_datasets}
\end{table*}

%\section{Impact of compression rate on performances}
%\label{appendix:variable_compression_rate}

%Figure \ref{fig:variable_compression_rate_solar} shows impact of compression rate on PISCO-Solar models on all datasets. Once again, the effect is a gradual decrease in performance as the compression rate increases. Notably, low-compression Solar models improve on the uncompressed Solar. We attribute this effect to the fine-tuning on the labels generated by Solar, which often include explanations on top of the raw question answer. Same results are shown on Figure \ref{fig:variable_compression_rate_mistral} for Mistral-backboned PISCO models.

%\begin{figure*}[!tb]
%    \centering
%    \includegraphics[width=\textwidth]{variable_compression_rate_solar.png} 
%    \caption{Performances (accuracy) of PISCO-Solar models with variable compression rates on all datasets. Dashed lines are no-compression baselines.}
%    \label{fig:variable_compression_rate_solar}
%\end{figure*}

%\begin{figure*}[!tb]
%    \centering
%    \includegraphics[width=\textwidth]{variable_compression_rate_mistral.png} 
%    \caption{Performances (accuracy) of PISCO-Mistral models with variable compression rates on all datasets. Dashed lines are no-compression baselines.}
%    \label{fig:variable_compression_rate_mistral}
%\end{figure*}

\section{PISCO with frozen decoder}
\label{appendix:frozen_decoder}
\begin{figure}[!tb]
    \centering
    \includegraphics[width=\columnwidth]{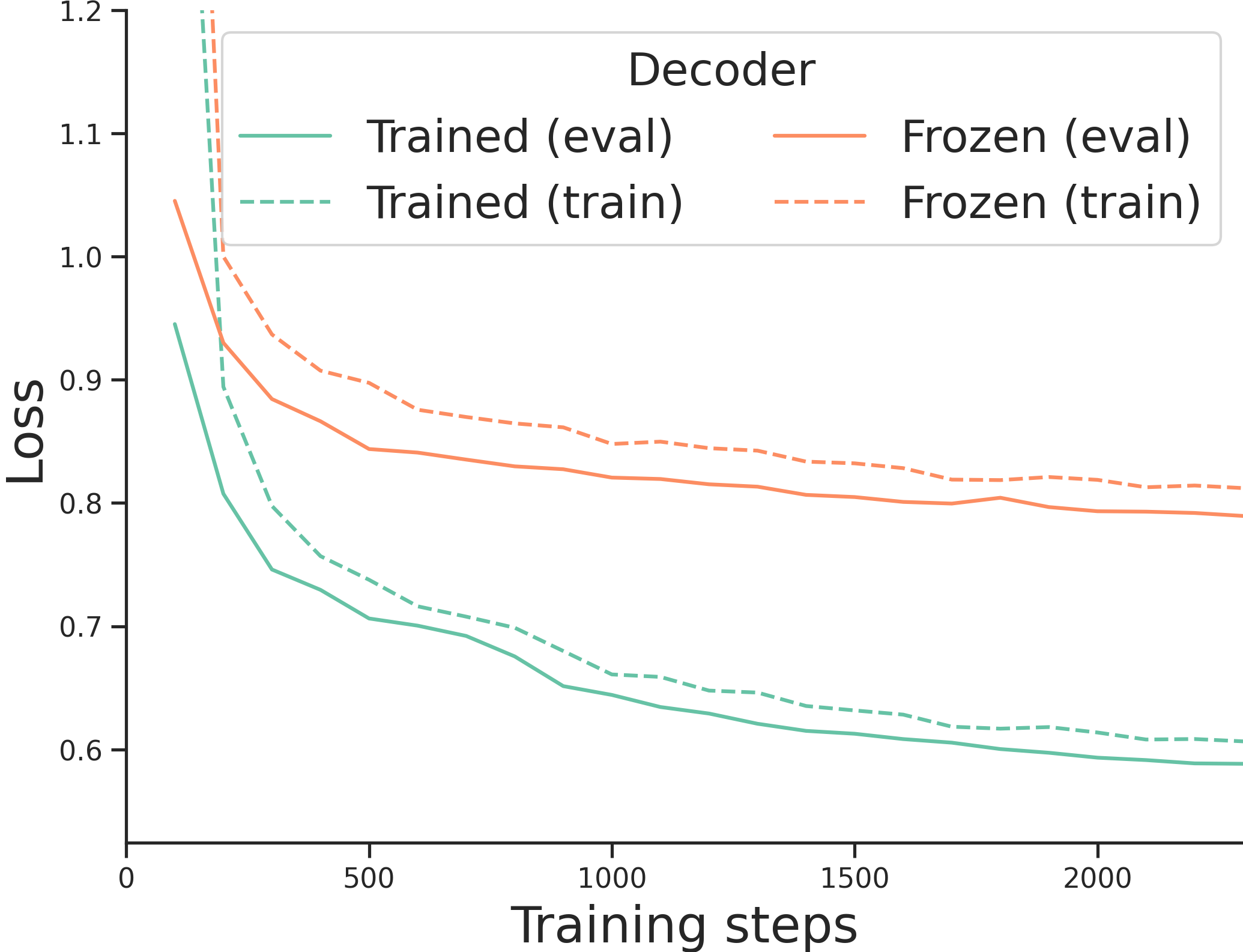} 
    \caption{Train and eval loss curves PISCO models with and without training the decoder.}
    \label{fig:loss_frozen}
\end{figure}

Freezing the decoder for compression models is appealing: it would enable to use compressed representations without any major change to the decoding pipeline of an existing system, as in \cite{cheng2024xrag}. To that end, we ran fine-tuning (with and without pre-training) of PISCO models with frozen decoder. Table \ref{table:frozen_decoder} shows the difference in performance is huge. In fact, a look at the loss curves \ref{fig:loss_frozen} seems to show that fitting only the compressor does not offer nearly enough flexibility for learning. 

\begin{table}
    \centering
    \begin{tabular}{lcc}
    \toprule
    Decoder & Trained & Frozen \\
    \midrule
    ASQA & 0.72 & 0.65 \\
    HotpotQA & 0.48 & 0.42 \\
    NQ & 0.65 & 0.72 \\
    TriviaQA & 0.90 & 0.87 \\
    POPQA & 0.66 & 0.51 \\
    \bottomrule
    \end{tabular}
    \caption{Performance (accuracy) on general domain with and without decoder training.}
    \label{table:frozen_decoder}
\end{table}

\section{Embeddings analysis}
\label{appendix:embedding_analysis}

\begin{figure}[!tb]
    \centering
    \includegraphics[width=\columnwidth]{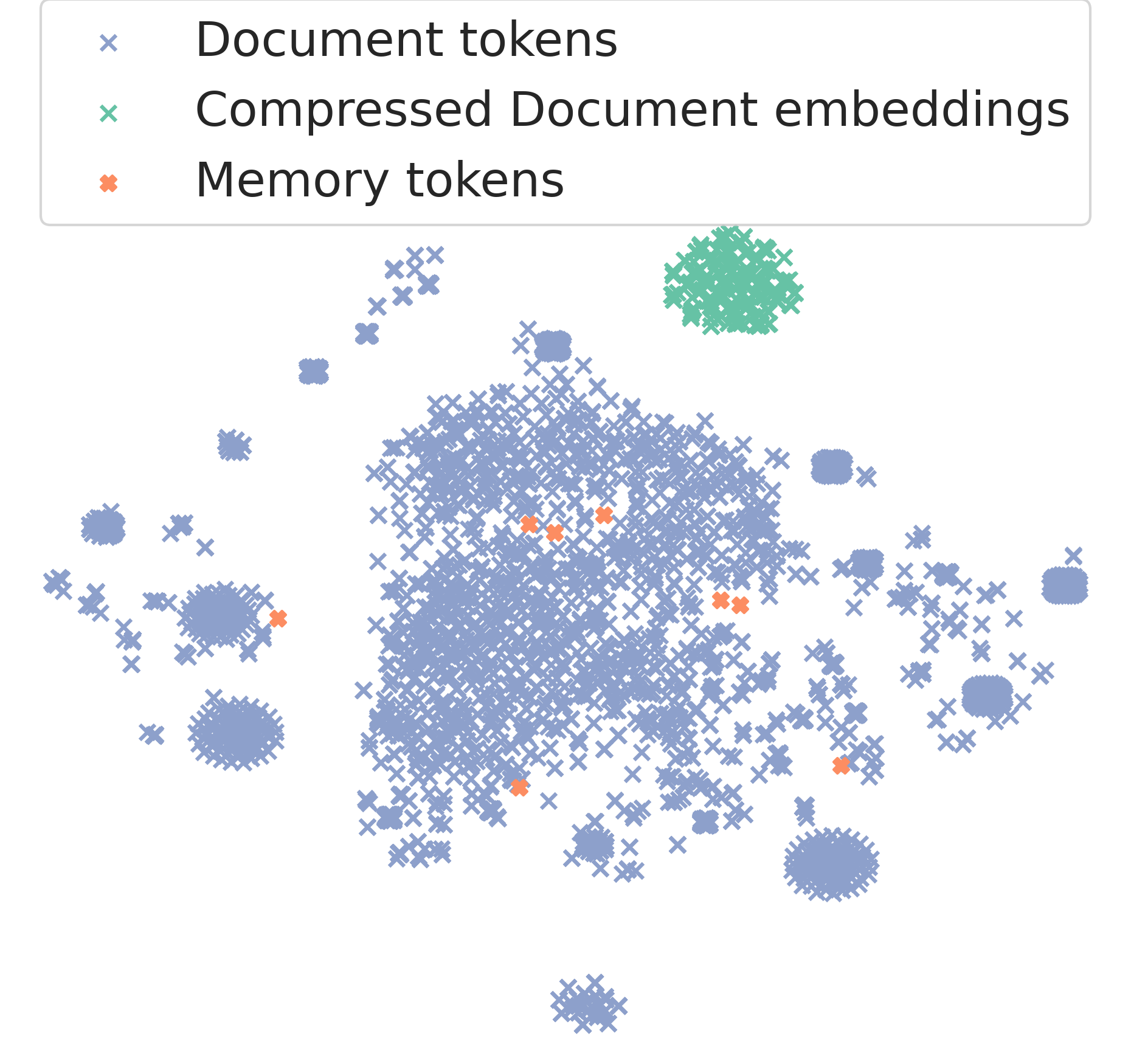} 
    \caption{t-SNE visualization of document tokens, memory tokens, and compressed document embeddings. The compressed embeddings lie outside the token distribution, supporting why freezing the decoder in a compression model is ineffective.}
    \label{fig:tnse_embedding}
\end{figure}

To better understand how information is compressed within the document embeddings, we apply the logit lens \cite{nostalgebraist2020logit} to each embedding. This allows us to identify the top 10 tokens by mapping the embeddings to logits space using the LLM head. An example of the results is provided in Table \ref{table:logits_attribution}. Most top tokens correspond or are close to some token in the compressed text. We also recover the spatial specialization shown on Figure \ref{fig:cosine_similary_embeddings}.

\begin{table*}
\begin{tabular}{| m{0.39\textwidth} | m{0.6\textwidth} |}
\hline
\textbf{Text} & \textbf{Logits attributions of the text embeddings} \\
\hline
\textcolor{algocolor5}{the} \textcolor{algocolor2}{Gardner-Harvey} \textcolor{algocolor2}{Paper} \textcolor{algocolor2}{company,} \textcolor{algocolor4}{installing} a very large \textcolor{algocolor2}{paper} \textcolor{algocolor0}{board} \textcolor{algocolor4}{machine,} \textcolor{algocolor0}{and} in 1916 \textcolor{algocolor6}{organized} \textcolor{algocolor5}{the} \textcolor{algocolor3}{Gardner} \textcolor{algocolor2}{Paper} \textcolor{algocolor0}{Board} \textcolor{algocolor2}{company} \textcolor{algocolor0}{and} \textcolor{algocolor6}{took} over \textcolor{algocolor5}{the} \textcolor{algocolor0}{old} \textcolor{algocolor0}{National} \textcolor{algocolor0}{Box} \textcolor{algocolor0}{Board} \textcolor{algocolor2}{company,} which was at that time in \textcolor{algocolor5}{the} \textcolor{algocolor0}{hands} of \textcolor{algocolor5}{the} \textcolor{algocolor0}{receiver.} All \textcolor{algocolor5}{three} of \textcolor{algocolor5}{these} \textcolor{algocolor2}{companies} \textcolor{algocolor5}{met} \textcolor{algocolor7}{with} \textcolor{algocolor5}{phenomenal} \textcolor{algocolor5}{success,} due to \textcolor{algocolor5}{the} \textcolor{algocolor3}{efforts} of Colin \textcolor{algocolor3}{Gardner,} who it is \textcolor{algocolor5}{conceded} was one of \textcolor{algocolor5}{the} most brilliant \textcolor{algocolor3}{business} men of his \textcolor{algocolor3}{day.} He was a \textcolor{algocolor3}{Republican,} but \textcolor{algocolor6}{took} \textcolor{algocolor1}{no} \textcolor{algocolor1}{active} part in \textcolor{algocolor1}{politics,} \textcolor{algocolor1}{nor} did he \textcolor{algocolor1}{care} \textcolor{algocolor1}{for} \textcolor{algocolor1}{fraternal} \textcolor{algocolor1}{connections.} \textcolor{algocolor7}{During} \textcolor{algocolor5}{the} \textcolor{algocolor7}{Civil} war he \textcolor{algocolor7}{served} \textcolor{algocolor7}{with} \textcolor{algocolor5}{the} \textcolor{algocolor7}{100-}  & \textbf{Token 0}: \textcolor{algocolor0}{receiver}, \textcolor{algocolor0}{hands}, giornata, \textcolor{algocolor0}{bo}, \textcolor{algocolor0}{boxes}, \textcolor{algocolor0}{national}, \textcolor{algocolor0}{box}, \textcolor{algocolor0}{board}, \textcolor{algocolor0}{old}, \textcolor{algocolor0}{nacional}\newline
\textbf{Token 1}: \textcolor{algocolor1}{cares}, \textcolor{algocolor1}{fr}, \textcolor{algocolor1}{connection}, neither, \textcolor{algocolor1}{cared}, \textcolor{algocolor1}{connections}, \textcolor{algocolor1}{actively}, \textcolor{algocolor1}{nor}, \textcolor{algocolor1}{politics}, \textcolor{algocolor1}{political}, \textcolor{algocolor1}{active}, \textcolor{algocolor1}{care}\newline
\textbf{Token 2}: \textcolor{algocolor2}{papers}, \textcolor{algocolor2}{paper}, \textcolor{algocolor2}{companies}, \textcolor{algocolor2}{company}, \textcolor{algocolor2}{harvey}, harold, \textcolor{algocolor2}{newspaper}, \textcolor{algocolor2}{pap}, \textcolor{algocolor2}{board}, \textcolor{algocolor2}{har}\newline
\textbf{Token 3}: \textcolor{algocolor3}{businesses}, \textcolor{algocolor3}{efforts}, \textcolor{algocolor3}{business}, \textcolor{algocolor3}{republican}, \textcolor{algocolor3}{gardens}, \textcolor{algocolor3}{republicans}, bright, \textcolor{algocolor3}{gard}, \textcolor{algocolor3}{garden}, \textcolor{algocolor3}{days}, \textcolor{algocolor3}{effort}, brains, \textcolor{algocolor3}{day}\newline
\textbf{Token 4}: \textcolor{algocolor4}{papers}, \textcolor{algocolor4}{machines}, \textcolor{algocolor4}{paper}, \textcolor{algocolor4}{company}, \textcolor{algocolor4}{companies}, \textcolor{algocolor4}{boards}, \textcolor{algocolor4}{pap}, \textcolor{algocolor4}{board}, \textcolor{algocolor4}{installation}, \textcolor{algocolor4}{machine}, \textcolor{algocolor4}{installed}, \textcolor{algocolor4}{install}\newline
\textbf{Token 5}: succeed, \textcolor{algocolor5}{met}, \textcolor{algocolor5}{efforts}, \textcolor{algocolor5}{three}, \textcolor{algocolor5}{phenomen}, \textcolor{algocolor5}{succeeded}, \textcolor{algocolor5}{success}, \textcolor{algocolor5}{meet}, ycne, \textcolor{algocolor5}{successful}, \textcolor{algocolor5}{effort}, achievements, meeting\newline
\textbf{Token 6}: \textcolor{algocolor6}{organ}, giornata, \textcolor{algocolor6}{company}, \textcolor{algocolor6}{paper}, \textcolor{algocolor6}{companies}, \textcolor{algocolor6}{boards}, \textcolor{algocolor6}{organiz}, \textcolor{algocolor6}{board}, \textcolor{algocolor6}{took}, organisation, take, \textcolor{algocolor6}{organized}, organization\newline
\textbf{Token 7}: \textcolor{algocolor7}{serv}, rera, \textcolor{algocolor7}{served}, volunteers, serving, servants, \textcolor{algocolor7}{with}, \textcolor{algocolor7}{during}, \textcolor{algocolor7}{civil}, \textcolor{algocolor7}{-}, \textcolor{algocolor7}{serves}, \textcolor{algocolor7}{service}, \textcolor{algocolor7}{serve}\newline
  \\
\hline
tenth of 21 \textcolor{algocolor4}{Franciscan} \textcolor{algocolor0}{missions} \textcolor{algocolor4}{built} \textcolor{algocolor0}{in} \textcolor{algocolor4}{upper} \textcolor{algocolor2}{California.} Soon \textcolor{algocolor6}{Yankee} \textcolor{algocolor6}{traders,} \textcolor{algocolor6}{tourists} and \textcolor{algocolor6}{health} seekers, followed by \textcolor{algocolor6}{wealthy} \textcolor{algocolor6}{Easterners} \textcolor{algocolor6}{settled} \textcolor{algocolor0}{in} \textcolor{algocolor0}{Santa} Barbara because of the \textcolor{algocolor0}{mild} \textcolor{algocolor0}{winters.} The \textcolor{algocolor0}{mixture} of \textcolor{algocolor0}{newcomers} and \textcolor{algocolor0}{Spanish} \textcolor{algocolor5}{descendants} has \textcolor{algocolor5}{shaped} the \textcolor{algocolor2}{area} for what it has become \textcolor{algocolor5}{today.} \textcolor{algocolor0}{Accommodations:} \textcolor{algocolor0}{Santa} Barbara \textcolor{algocolor2}{boasts} \textcolor{algocolor1}{over} 90 \textcolor{algocolor3}{motels} and \textcolor{algocolor3}{hotels,} \textcolor{algocolor1}{plus} \textcolor{algocolor3}{numerous} \textcolor{algocolor3}{bed} and breakfast inns, all of which provide \textcolor{algocolor1}{over} 4,500 \textcolor{algocolor1}{rooms} \textcolor{algocolor1}{from} the modest to the \textcolor{algocolor1}{mos6t} luxurious for \textcolor{algocolor1}{both} \textcolor{algocolor7}{business} \textcolor{algocolor6}{travelers} and \textcolor{algocolor7}{tour}  & \textbf{Token 0}: attracted, \textcolor{algocolor0}{accommod}, attract, \textcolor{algocolor0}{mixture}, \textcolor{algocolor0}{spanish}, \textcolor{algocolor0}{mi}, \textcolor{algocolor0}{santa}, \textcolor{algocolor0}{new}, \textcolor{algocolor0}{mild}, \textcolor{algocolor0}{win}, \textcolor{algocolor0}{winter}, kennis\newline
\textbf{Token 1}: \textcolor{algocolor1}{accommod}, \textcolor{algocolor1}{mos}, \textcolor{algocolor1}{from}, \textcolor{algocolor1}{plus}, \textcolor{algocolor1}{room}, \textcolor{algocolor1}{over}, \textcolor{algocolor1}{both}, \textcolor{algocolor1}{rooms}, thousand, kennis\newline
\textbf{Token 2}: \textcolor{algocolor2}{accommod}, \textcolor{algocolor2}{california}, \textcolor{algocolor2}{bo}, biologie, \textcolor{algocolor2}{santa}, \textcolor{algocolor2}{area}, \textcolor{algocolor2}{acc}, plaat, områ, \textcolor{algocolor2}{accom}, \textcolor{algocolor2}{over}, \textcolor{algocolor2}{accommodate}, \textcolor{algocolor2}{área}, \textcolor{algocolor2}{accommodation}, kennis\newline
\textbf{Token 3}: \textcolor{algocolor3}{accommod}, countless, \textcolor{algocolor3}{mot}, \textcolor{algocolor3}{hotels}, \textcolor{algocolor3}{beds}, \textcolor{algocolor3}{plus}, \textcolor{algocolor3}{hotel}, \textcolor{algocolor3}{bed}, \textcolor{algocolor3}{numerous}, \textcolor{algocolor3}{accommodation}\newline
\textbf{Token 4}: lower, \textcolor{algocolor4}{california}, building, \textcolor{algocolor4}{built}, \textcolor{algocolor4}{upper}, \textcolor{algocolor4}{missions}, \textcolor{algocolor4}{mission}, \textcolor{algocolor4}{up}, \textcolor{algocolor4}{build}, \textcolor{algocolor4}{francis}\newline
\textbf{Token 5}: \textcolor{algocolor5}{shape}, \textcolor{algocolor5}{today}, \textcolor{algocolor5}{area}, \textcolor{algocolor5}{spanish}, \textcolor{algocolor5}{descend}, \textcolor{algocolor5}{new}, \textcolor{algocolor5}{span}, \textcolor{algocolor5}{sha}, \textcolor{algocolor5}{accom}, \textcolor{algocolor5}{areas}, \textcolor{algocolor5}{accommodation}, \textcolor{algocolor5}{shapes}, \textcolor{algocolor5}{shaped}\newline
\textbf{Token 6}: attracted, \textcolor{algocolor6}{settled}, \textcolor{algocolor6}{yan}, \textcolor{algocolor6}{sett}, \textcolor{algocolor6}{health}, \textcolor{algocolor6}{healthy}, \textcolor{algocolor6}{settlement}, \textcolor{algocolor6}{healthcare}, \textcolor{algocolor6}{eastern}, \textcolor{algocolor6}{settle}, \textcolor{algocolor6}{wealthy}, \textcolor{algocolor6}{traders}, \textcolor{algocolor6}{tourists}, kennis\newline
\textbf{Token 7}: \textcolor{algocolor7}{businesses}, \textcolor{algocolor7}{travel}, luxury, \textcolor{algocolor7}{tourist}, \textcolor{algocolor7}{business}, \textcolor{algocolor7}{tour}, \textcolor{algocolor7}{tours}, \textcolor{algocolor7}{both}, tournament, \textcolor{algocolor7}{tourists}, alike\newline
 \\
 \hline
\end{tabular} 
\caption{Text and the logits attribution of its memory embeddings: for each memory embedding, we compute the top-10 tokens using the head matrix of the decoder. Almost all top tokens correspond to a token in the text. Each memory embedding puts more emphasis on some part of the text.}
\label{table:logits_attribution}
\end{table*}

\begin{table}[!tb]
    \centering
    \begin{tabular}{|llll|}
    \hline
    \multicolumn{4}{|c|}{Pretraining Mixtures} \\
    \hline \hline
     AE & TC & KBTC & multi-KBTC \\ \hline
     0.5 & 0.5 & 0. & 0. \\ \hline
     0.25 & 0.25 & 0.5 & 0. \\ \hline
     0. & 0.5 & 0.5 & 0. \\ \hline
     0. & 0.25 & 0.75 & 0. \\ \hline
     0. & 0.25 & 0.25 & 0.5 \\
    \hline
    \end{tabular}
    \caption{Tested pretraining configurations. All were run with compression rate of 16 and 128.}
    \label{table:pretraining_mixtures}
\end{table}

\section{Attempted pretraining tasks}
\label{appendix:pretraining_tasks}

As described in \S \ref{subsection:pisco_is_optimal}, our initial approach focused on designing more complex pretraining tasks. Pretraining followed the lines and configurations of \cite{rau2024context}. The pretraining task we implemented consisted in:
\begin{itemize}
    \item Auto-encoding (AE): the compressed representation of a single document as well as a task-specific token <AE> is prompted to the decoder during pretraining: the labels is the plain text document.
    \item Text Continuation (TC): as for general LM training, this task prompts the decoder with the compressed representation of the document and its task is to generate the following text.
    \item Keyword-Based Text Continuation (KBTC): A potential concern --especially since auto-encoding works so well even with high compression rates-- was that accessing information within the middle of texts while working on their compressed representations was difficult. To address this, the keyword-based text continuation task prompted the decoder with the compressed document as well as a small sequence (the keyword) extracted randomly from the compressed text. Its target was to generate what followed this keyword in the text. Here also it is possible to reach a very high Rouge-L in string reconstruction, with no effect on final QA performance. This showed that lost-in-the-middle effect was not really a concern within compressed documents.
    \item Multi-document Keyword-Based Text Continuation (multi-KBTC) is identical to Keyword-Based Text Continuation except that multiple encoded documents are prompted to the decoder at once. The motivation here was to address a potential between-document lost-in-the-middle effect.
\end{itemize}

In all cases, the loss was the cross-entropy loss on the target generation. The configurations analysed in the paper consist of mixtures of these different tasks and are described on Table \ref{table:pretraining_mixtures}.

We also tested formulating these tasks as instruction tasks with corresponding prompts as in \cite{cheng2024xrag}, without further success.

\section{String Normalization for metric computation}
\label{appendix:normalize}

To measure accuracy, F1 score or recall between a ground truth label and a prediction, we check that the normalized label is included in the normalized prediction. When multiple labels are possible, we take maximum values across the available labels. Normalization consists in:
\begin{itemize}
    \item Converting the string to lowercase
    \item Removing punctuation
    \item Removing articles: “a”, “an”, “the”
    \item Standardizing word splits by replacing multiple spaces and line returns with a single space
\end{itemize}

\end{document}